\documentclass[10pt,twocolumn,letterpaper]{article}
\usepackage{authblk}

\usepackage[pagenumbers]{iccv}      %

\usepackage{xcolor}
\usepackage{multicol,multirow}
\usepackage{overpic}
\usepackage{float}
\usepackage{wrapfig}
\usepackage{amsmath}
\usepackage{graphicx}
\usepackage{amssymb}
\usepackage{booktabs}
\usepackage{afterpage}
\usepackage{pifont}%
\usepackage{wrapfig}
\usepackage{makecell}
\usepackage{dsfont}
\usepackage{nccmath}
\usepackage{mathtools}
\usepackage{nicefrac}
\usepackage{listings}
\usepackage[accsupp]{axessibility}

\floatstyle{plain}
\newfloat{verbatimcaption}{thp}{lop}
\floatname{verbatimcaption}{Listing}

\newcommand{\methodname}[0]{TriDi}
\newcommand{\methodnetwork}[0]{\mathrm{TriDi}_\psi }
\newcommand{\methodsingleOI}[0]{s-TriDi-OI }
\newcommand{\methodsingleHI}[0]{s-TriDi-HI }
\newcommand{\methodsingleHO}[0]{s-TriDi-HO }

\newcommand{\h}[0]{\mathcal{H} }
\newcommand{\hpose}[0]{\mathbf{\theta}_\h }
\newcommand{\hidentity}[0]{\mathbf{\beta}_\h }

\newcommand{\htemplatev}[0]{\mathbf{V}_\h }

\newcommand{\hglobal}[0]{\mathbf{g}_\h }
\newcommand{\htime}[0]{t^\h}

\newcommand{\obj}[0]{\mathcal{O} }

\newcommand{\otemplatev}[0]{\mathbf{V}_\obj }

\newcommand{\objglobal}[0]{\mathbf{g}_\obj }
\newcommand{\objfeatures}[0]{\mathbf{f}_\obj }
\newcommand{\objconditioning}[0]{\mathcal{C}_\obj }
\newcommand{\objonehot}[0]{\mathbf{y}_\obj }
\newcommand{\objtime}[0]{t^\obj} 

\newcommand{\inter}[0]{\mathcal{I}}
\newcommand{\interlabel}[0]{T_{\inter}}
\newcommand{\interlatent}[0]{\mathbf{z}_{\inter}}
\newcommand{\interheat}[0]{\phi_{\inter}}

\newcommand{\intertime}[0]{t^{\inter}}

\newcommand{\predhpose}[0]{\hat{\mathbf{\theta}}_\h }
\newcommand{\predhidentity}[0]{\hat{\mathbf{\beta} }_\h}
\newcommand{\predhglobal}[0]{\hat{\mathbf{g}}_\h }
\newcommand{\predobjglobal}[0]{\hat{\mathbf{g}}_\obj }
\newcommand{\predinterlatent}[0]{\hat{\mathbf{z}}_{\inter}}
\newcommand{\predinterheat}[0]{{\hat{\phi}_{\inter}}}

\newcommand{\predotemplatev}[0]{\hat{\mathbf{V}}_\obj }
\newcommand{\predhtemplatev}[0]{\hat{\mathbf{V}}_\h }

\newcommand{\first}[1]{\mathbf{#1}}
\newcommand{\second}[1]{\underline{#1}}

\definecolor{input_col}{RGB}{13, 103, 152}
\definecolor{pred_col}{RGB}{16, 122, 19}

\definecolor{cond_col}{RGB}{204, 204, 0}
\definecolor{t_col}{RGB}{154,1,150}

\newcommand{\colin}[1]{\textcolor{input_col}{#1}}
\newcommand{\colout}[1]{\textcolor{pred_col}{#1}}

\newcommand{\genhi}[0]{$\colout{\h},\colout{\inter}|\colin{\obj}$}
\newcommand{\genoi}[0]{$\colout{\obj},\colout{\inter}|\colin{\h}$}

\newcommand{\genh}[0]{$\colout{\h}|\colin{\obj},\colin{\inter}$}
\newcommand{\geno}[0]{$\colout{\obj}|\colin{\h},\colin{\inter}$}
\newcommand{\geni}[0]{$\colout{\inter}|\colin{\h},\colin{\obj}$}
\newcommand{\genho}[0]{$\colout{\h},\colout{\obj}|\colin{\inter}$}
\newcommand{\genhoi}[0]{$\colout{\h},\colout{\obj},\colout{\inter}$}

\definecolor{iccvblue}{rgb}{0.21,0.49,0.74}
\usepackage[pagebackref,breaklinks,colorlinks,allcolors=iccvblue]{hyperref}

\title{TriDi: Trilateral Diffusion of 3D Humans, Objects, and Interactions}

\makeatletter
\renewcommand\AB@affilsepx{ , \protect\Affilfont}
\makeatother

\author[1,2]{Ilya A. Petrov}
\author[3,4]{Riccardo Marin}
\author[1,5]{Julian Chibane}
\author[1,2,5]{Gerard Pons-Moll}

\affil[1]{\small University of T\"ubingen, Germany}
\makeatletter
\renewcommand\AB@affilsepx{\protect, \\\protect\Affilfont}
\makeatother
\affil[2]{\small T\"ubingen AI Center, Germany}
\makeatletter
\renewcommand\AB@affilsepx{\protect, \protect\Affilfont}
\makeatother
\affil[3]{\small Technical University of Munich, Germany}
\makeatletter
\renewcommand\AB@affilsepx{\protect, \\\protect\Affilfont}
\makeatother
\affil[4]{\small Munich Center for Machine Learning, Germany}
\makeatletter
\renewcommand\AB@affilsepx{\protect\\\protect\Affilfont}
\makeatother
\affil[5]{\small Max Planck Institute for Informatics, Saarland Informatics Campus, Germany}
\affil[ ]{\tt\small \{i.petrov, gerard.pons-moll\}@uni-tuebingen.de}
\affil[ ]{\tt\small riccardo.marin@tum.de, jchibane@mpi-inf.mpg.de}

\begin{document}
\twocolumn[{%
\renewcommand\twocolumn[1][]{#1}%
\maketitle
\begin{center}
    \centering
    \captionsetup{type=figure}
    \vspace{-15pt}
    \includegraphics[width=0.95\linewidth]{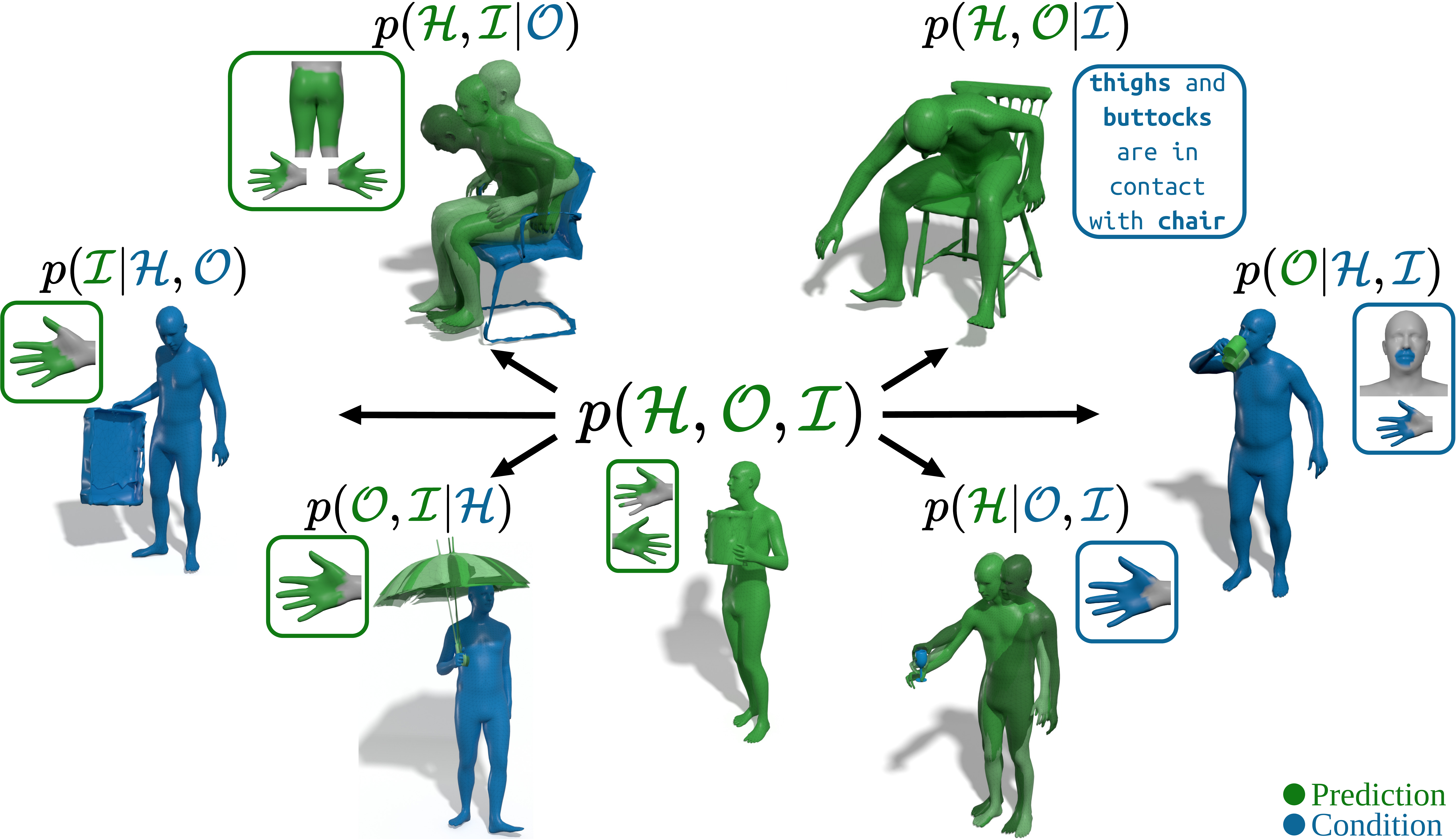}
    \vspace{-9pt}
    \caption{\textbf{{\methodname}}. We present {\methodname}, the first joint probabilistic model of human pose ($\h$), object ($\obj$) and human-object interaction ($\inter$). 
    The joint model unifies these three modalities, capturing mutual dependencies between them, and allows for sampling in \emph{seven} conditioning configurations, covering the use cases treated in isolation by previous works. 
    The colors on the image encode {\color{pred_col}prediction} and {\color{input_col}condition}. 
    }
    \label{fig:teaser}
    \vspace{-4pt}
\end{center}%
}]

\begin{abstract}
\vspace{-1pt}
Modeling 3D human-object interaction (HOI) is a problem of great interest for computer vision and a key enabler for virtual and mixed-reality applications. 
Existing methods work in a one-way direction: some recover plausible human interactions conditioned on a 3D object; others recover the object pose conditioned on a human pose. 
Instead, we provide the first unified model - \emph{{\methodname}} which 
works in any direction.
Concretely, we generate Human, Object, and Interaction modalities simultaneously with a new three-way diffusion process, allowing to model \emph{seven} distributions with one network.
We implement {\methodname} as a transformer attending to the various modalities' tokens, thereby discovering conditional relations between them. 
The user can control the interaction either as a text description of HOI or a contact map. 
We embed these two representations into a shared latent space, combining the practicality of text descriptions with the expressiveness of contact maps. 
Using a single network, {\methodname} unifies all the special cases of prior work and extends to new ones, modeling a family of \emph{seven} distributions. 
Remarkably, despite using a single model, {\methodname} generated samples surpass one-way specialized baselines on GRAB and BEHAVE in terms of both qualitative and quantitative metrics, and demonstrating better diversity. 
We show the applicability of {\methodname} to scene population, generating objects for human-contact datasets, and generalization to unseen object geometry.
The project page is available at: \href{https://virtualhumans.mpi-inf.mpg.de/tridi/}{https://virtualhumans.mpi-inf.mpg.de/tridi/}.
\vspace{-3pt}
\end{abstract}

\section{Introduction}
\label{sec:introduction}

Humans constantly interact with objects around them -- they lean on tables, carry backpacks, or touch keyboards. Different \emph{objects} afford different kinds of \emph{human poses}, and vice-versa, different poses support only certain types of objects. 
Furthermore, given a  \emph{human} and an \emph{object}, many \emph{interactions} are possible. 
For example, we can sit on a chair, lift it, push it, or carry it, and each interaction will require different contacts. 
We argue that a comprehensive model should capture such interplay of objects, humans, and interactions, regardless of the modality considered as an input condition. 
Such a joint model is much more flexible than one-way models, giving rise to many applications: generating humans that fit a given object, objects that fit a human pose, unconditional generation, or even extending the annotations of existing 3D human-contact datasets with objects. 
This versatility is needed in such applications as content creation, AR/VR, ergonomics, and manufacturing. 

However, existing works have modeled human-object interaction as the posteriors of human given the object~\cite{zhang2022couch, kulkarni2023nifty, li2023object,taheri2022goal, wu2022saga} or object given human~\cite{petrov2023popup, yi2022humanaware}.
Following this paradigm requires a tailored model for each conditioning case and, thus, a specialized design choices, training procedure, and architecture.
Such an approach is impractical and difficult to scale. 
Instead of modeling each individual conditional distribution, we shift this paradigm and design a single compact architecture that models the joint and conditional distributions of \emph{human}, \emph{object}, and \emph{interactions}.
By design, we can sample from a joint unconditional distribution of humans, objects, and interactions, as well as from all possible conditional combinations.

We propose \textbf{{\methodname}}, a unified 3D human-object interaction model capturing the joint distribution of \emph{humans}, \emph{objects}, and \emph{interactions}. {\methodname} produces samples from every conditional distribution arising from the combination of three in addition to the joint distribution, giving rise to $2^3-1=7$ possible modes of operation, see Fig.~\ref{fig:teaser}.

{\methodname} performs a three-way diffusion building on the UniDiffuser paradigm \cite{bao2023one}, implemented through token-wise attention, enabling to capture fine-grained relations. 
Since most interactions imply contact, prior work represents interaction through body contact maps \cite{hassan2021populating, tripathi2023deco}. In contrast to text prompts, this representation is difficult for users to control. 
Hence, we propose to unify textual descriptions and body contact maps by a joint embedding space. 
This results in a novel representation that is useful to guide the model and intuitive to the user.
To double the effective training data and remove right-handed biases, we augment the HOI by exploiting the left-right symmetry of the interactions.

We demonstrate the flexibility of {\methodname}, which nicely encapsulates uni-directional methods published in different papers using a single network. 
Beyond savings in terms of model size and ease of use, {\methodname} surpasses uni-directional baselines tailored to specific conditioning cases. 
Moreover, {\methodname} performs on par or better than the same model trained on one-way conditional tasks (e.g., generating human and interactions conditioned on the object), demonstrating the effectiveness of the joint modeling.    
{\methodname} is general and capable of synthesizing a static 3D HOI starting from different inputs, covering all the previous works' use cases plus new ones (Fig.~\ref{fig:teaser}). 
We demonstrate how {\methodname} can populate scenes with realistic interactions, generalize to novel geometry, and open new applications such as generating objects that fit observed humans and interactions in images.

In summary, our contributions are:
\begin{itemize}
    \item We formulate {\methodname}, the first joint model for $P(\h,\obj,\inter)$, modeling it as the three-variable joint distribution and covering a total of \emph{7} modes of operation, rendering prior works as special use cases of our model.  
    \item We propose a novel representation of interaction by jointly embedding body contact maps and textual descriptions, resulting in intuitive control for the user while providing detailed guidance to the model. 
    \item We will release our code, providing the community with a tool for scene population, generation from partial observations, and other tasks that involve 3D HOI.

\end{itemize}

\section{Related Work}
\label{sec:related_work}
\paragraph{From object to human.} 
Modeling 3D HOI from the objects has been studied from diverse perspectives.
At a macro scale, studying humans in the context of 3D scenes is prominent~\cite{hassan2019resolving, hassan2021populating, zhang2020generating, zhang2020place, huang2022capturing, mir2023generating, wang2021synthesizing, Wang_2021_CVPR, yi2024generating}. 
These works are instrumental for downstream tasks like synthetic dataset generation~\cite{black2023bedlam, patel2021agora, xie2024procigen, jiang2024scaling}.
Dynamic motions in scenes can be conditioned by object's 3D location~\cite{chao2021learning, hassan2021stochastic}, control points~\cite{zhang2022couch}, milestones~\cite{pi2023hierarchical, li2023object}, physical properties~\cite{zhang2024force, Sang_2025_CVPR,sang2025twosquared}, or text descriptions~\cite{li2024controllable, song2024hoianimator}.
Such a high-level perspective on HOI should be complemented by modeling interactions on an object level.
Hence, ~\cite{wu2022saga, taheri2022goal} synthesize the motion towards a static object, and \cite{braun2024physically, li2024task} focus on manipulation interactions. 
These works consider temporal sequences, which are demanding to capture, thus limiting the scaling beyond the settings seen at training time.
Synthesizing hand-object interactions presents several challenges~\cite{feix2014analysisa, feix2016grasp}, which originated specialized methods~\cite{christen2022dgrasp, liu2009dextrous, ye2012synthesis, jiang2021hand, zhang2021manipnet, wang2023dexgraspnet, liu2021synthesizing}.
Producing accurate prediction raised the demand for hand-object refinement~\cite{luo2024physics, zhou2024gears, taheri2024grip, zhou2022toch}, but those methods are limited by smaller objects and hand-held interactions.
{\methodname} works with single frames, models contact beyond the hands, and thus supports human synthesis involving diverse objects.

\paragraph{From human to object.} 
Reasoning about objects from humans is a less explored direction, despite the applicability in AR/VR, where humans often interact with objects without a physical counterpart. 
~\cite{yi2023mime, ye2022scene} generate scenes satisfying the observed human motion.
Object Pop-up~\cite{petrov2023popup} regresses an object position from a 3D human point cloud, disregarding the uncertainty behind this ill-posed task. 
An interesting self-supervised approach regresses the heatmap for plausible object center location \cite{han2023chorus}, while the follow-up work~\cite{kim2024coma} studies objects' affordances. 
{\methodname} models a joint distribution of HOI, naturally allowing for the uncertainty in predictions while retaining downstream applications.

\paragraph{Contacts modeling.}
Contact is the physical medium of many human-object interactions. 
In practice, contact maps are a good proxy to promote realism \cite{hassan2021populating, nam2024joint, wu2022saga}; however, their capture is often complicated by manual annotation \cite{tripathi2023deco, Cseke_2025_CVPR} or the need for specialized hardware \cite{brahmbhatt2019contactdb}.
Contact is represented in a range of ways, e.g., as distances \cite{diller2024cghoi}, proximity \cite{zhang2020place}, or maps on the body \cite{tripathi2023deco} and the object \cite{fan2023arctic, Dwivedi_2025_CVPR}.
Contact is often modeled on the hands \cite{brahmbhatt2019contactdb, brahmbhatt2020contactpose, grady2021contactopt}, with recent works considering the full body \cite{hassan2021populating, tripathi2023deco, Dwivedi_2025_CVPR}. 
An alternative is to represent the interaction through text \cite{diller2024cghoi, yi2024generating, song2024hoianimator}. 
While more interpretable and controllable, this representation limits the possibility of spatial reasoning for the methods.
In our work, we combine text and contact maps in a shared latent space, inheriting the advantage of both.

\paragraph{Joint modeling.}
A number of works focus on reconstructing interactions with single objects external data such as images \cite{wang2022reconstructing, weng2021holistic, zhang2020perceiving, xie2022chore, nam2024joint, xie2024procigen, chen2023detecting, tripathi2023deco, yang2024lemon, liu2024primitive}, videos \cite{yi2022humanaware, xie2023visibility, zhao2024imhoi}, and multi-view recording setups \cite{zhang2023neuraldome, jiang2022neuralhofusion}.
These works are backed by recent HOI data collections \cite{Liu_2022_CVPR, bhatnagar2022behave, fan2023arctic, huang2022intercap, jiang2023full, yang2024fhoi}. 
Modeling hand-object interactions jointly requires tailored methods~\cite{karunratanakul2020grasping}.
FLEX~\cite{tendulkar2023flex} combines grasp with full-body generation to fit HOI samples in the scene constraints.
IMoS~\cite{ghosh2023imos} and InterDiff~\cite{xu2023interdiff} start from past observations to forecast the continuation of a 3D HOI sequence. 
CG-HOI~\cite{diller2024cghoi} synthesizes human and object motion from text, training only on one dataset at a time.
These methods rely on strong conditioning: temporal HOI sequence, deterministic future, and text. 
Using single-frame data, although challenging and ambiguous, is more general and scalable.
Finally, we find it exciting to mention recent works in nascent fields: compositional shape generation including human and object~\cite{kim2023ncho, chen2024comboverse, cheng2023progressive3d, dai2024interfusion}, and modeling multiple human and object interactions~\cite{yin2023hi4d, shapovalov2023replay, muller2023generative, zhang2024hoim3, kim2024parahome}, suggesting more complex synergies in HOI. 
{\methodname} models HOI jointly, covering all the use cases of previous works tailored to the specific conditioning.

\section{Background}
\label{sec:background}
\paragraph{Probabilistic Diffusion.}

A Diffusion process \cite{sohl2015deep, ho2020denoising} is divided into a forward process that progressively noises the original data sample $\mathbf{z}_0$, and a backward process that recovers the sample $\mathbf{z}_0$ from the noise using a learned model.

Formally, the forward process follows a Markov chain of $T$ steps; it produces a series of time-dependent distributions $q(\mathbf{z}_t | \mathbf{z}_{t-1})$: $q(\mathbf{z}_{1:T} | \mathbf{z}_0) = \prod_{t=1}^{T}{q(\mathbf{z}_t | \mathbf{z}_{t-1})}$. At every timestamp, we inject noise into the distribution until the final $\mathbf{z}_T$ converges to a sample from $\mathcal{N}(\mathbf{0},\mathbf{I})$. Let $\beta_0=0$, and $\beta_t \in (0,1)$:
\begin{equation}
    \begin{aligned}
        q(\mathbf{z}_t | \mathbf{z}_{t-1}) & = \mathcal{N}(
          \mathbf{z}_t; 
          \sqrt{1 - \beta_t} \mathbf{z}_{t-1},
          \beta_t \mathbf{I}
        ).
    \end{aligned}
\end{equation}

We follow the formulation of Denoising Diffusion Probabilistic Model (DDPM) \cite{ho2020denoising} to obtain a closed-form expression for $\mathbf{z}_t$ (formulation is provided in the Sup. Mat.).

The inference is then performed by reversing the process, starting from $\mathbf{z}_T \sim \mathcal{N}(0, \mathbf{I})$ and recovering samples from the original distribution. 
Instead of recovering the added noise $\epsilon$ for each timestep, we follow the formulation of \cite{ramesh2022hierarchical} and recover the original sample $\mathbf{z}_0$. 
To achieve this, we parametrize the reverse process by a denoising neural network $\mathcal{D}_\psi$ that is trained to recover the original sample $\mathbf{z}_0$ from the noised sample $\mathbf{z}_t$ at timestep $t$ given the condition $c$. Defining for brevity 
$\mathbb{E}_p \equiv \mathbb{E}_{\mathbf{z}_0 \sim p_{data}}$, 
$\mathbb{E}_t \equiv \mathbb{E}_{t \sim \mathcal{U}\{0,...,T\}}$, and 
$\mathbb{E}_q \equiv \mathbb{E}_{\mathbf{z}_t \sim q(\mathbf{z}_t | \mathbf{z}_0)}$ 
we obtain the training objective (inference formulation is provided in the Sup. Mat.):

\begin{equation}
    \begin{aligned}
        \min_\psi  
           \mathbb{E}_p\, \mathbb{E}_t\, \mathbb{E}_q\, \|\mathcal{D}_\psi(\mathbf{z}_t; c, t) - \mathbf{z}_0\|).
    \end{aligned}
    \label{eqn:objective}
\end{equation}

\paragraph{Multimodal diffusion.}
While the previous formulation handles the generation of a single modality, data often constitutes a composition of multiple modalities, e.g., $\mathbf{z}_0 = (\mathbf{x}_0, \mathbf{y}_0) \sim p(x, y)$. Hence, we are naturally interested in modeling this joint distribution together with the marginals $p(y)$ and $p(x)$, as well as conditional ones $p(x|y)$ and  $p(y|x)$. UniDiffuser~\cite{bao2023one} proposes a network $\mathcal{D}_\psi(\mathbf{x}_{t^x}, \mathbf{y}_{t^y}; t^x, t^y)$ dedicated to recovering $\mathbf{z}_0$ given a noisy sample from the joint distribution.

Adapting the definitions from Eq.~\ref{eqn:objective} to two modalities: 
$\mathbb{E}_p \equiv \mathbb{E}_{(\mathbf{x}_0, \mathbf{y}_0) \sim p(x,y)}$, 
$\mathbb{E}_t \equiv \mathbb{E}_{(t^x, t^y) \sim \mathcal{U}\{0,...,T\}^2}$, and
$\mathbb{E}_q \equiv \mathbb{E}_{\mathbf{x}_{t^x} \sim q(\mathbf{x}_{t^x} | \mathbf{x}_0),
        \mathbf{y}_{t^y} \sim q(\mathbf{y}_{t^y} | \mathbf{y}_0)}$ 
we obtain the following training objective:
\begin{equation}
    \begin{aligned}
        \min_\psi 
            \mathbb{E}_p\, \mathbb{E}_t\, \mathbb{E}_q\, \|\mathcal{D}_\psi(\mathbf{x}_{t^x}, \mathbf{y}_{t^y}; t^x, t^y) - (\mathbf{x}_0, \mathbf{y}_0)\|.
    \label{eqn:unidiffuser}
    \end{aligned}
\end{equation}

The benefit of minimizing the objective in Eq.~\ref{eqn:unidiffuser} is that the resulting network captures all the desired distributions. Namely, setting $t^y = T$ allows to model the marginal distribution $p(x)$, on the other hand, $t^y = 0$ corresponds to conditional distribution $p(x | y)$. we note that in its original formulation, UniDiffuser is designed to consider text and images as two diffusion modalities.

\begin{figure*}[!ht]
    \centering
    \begin{overpic}[trim=0cm 0cm 0cm 0cm,clip, width=0.95\linewidth]{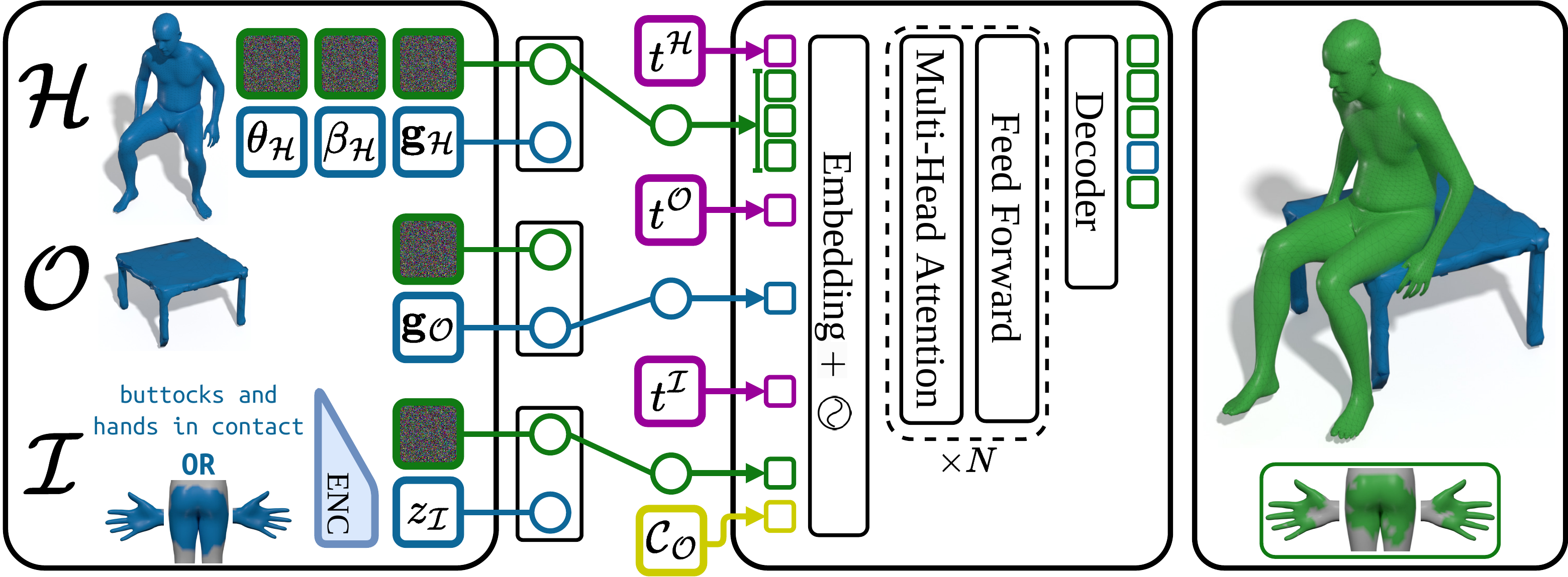}
        \put(11,38.5){\LARGE Input}
        \put(56,38.5){\LARGE Model}
        \put(84,38.5){\LARGE Output}
    \end{overpic}
    \vspace{-0.15cm}
    \caption{\textbf{{\methodname} Overview.}
        {\methodname} is a Trilateral Diffusion for Human $\h$ (pose $\hpose$, identity $\hidentity$, and 6-DoF global pose $\hglobal$), Object $\obj$ (6-DoF global pose $\objglobal$) and Interaction $\inter$ (Contact-Text latent $\interlatent$). 
        In this figure the model is configured to sample $p({\color{pred_col}{\h}},{\color{pred_col}{\inter}} | {\color{input_col}{\obj}})$. One of the \emph{seven} operating modes is chosen by adjusting the \textbf{\textcolor{t_col}{timestamp}} to be $0$ for a \textbf{\textcolor{input_col}{given condition}} ($\objtime$ above) and $T$ for the \textbf{\textcolor{pred_col}{desired prediction}} ($\htime$ and $\intertime$ above), and supplying an \textbf{\textcolor{cond_col}{object class condition}} (``Table'' above).}
    \label{fig:pipeline}
    \vspace{-0.5cm}
\end{figure*}

\section{Method} 
\label{sec:method}
\paragraph{Overview.} Our goal is to model the three-variable joint distribution of Human $\h$, Object $\obj$, and Interaction $\inter$, taking as input only the object class with a canonical representation and, optionally, conditions from the three modalities.
Previous works focus on one-way cases, with fixed conditional modality, e.g., human from an object, ($P(\h|\obj)$, \cite{li2023object}) or human and object from a text ($P(\h,\obj|\inter)$, \cite{diller2024cghoi}).
In contrast, we want to model $P(\h,\obj,\inter)$, providing \emph{a unified model for Human-Object Interaction}. 
To achieve this, we introduce {\methodname}, a transformer based model that operates on tokenized representations of $\h$, $\obj$, and $\inter$ (an overview is presented in Fig.\ref{fig:pipeline}).
The following sections define the representations for HOI (Sec.~\ref{sec:method_modalities}), introduce our trilateral diffusion formulation (Sec.~\ref{sec:method_diffusion}), and discuss training details (Sec.~\ref{sec:method_training}).

\subsection{Modalities representations}
\label{sec:method_modalities}
\paragraph{Human and object.} Following SMPL+H body model \cite{loper2015smpl, romero2017mano} we decompose the human as:
\vspace{-2.5pt}
\begin{equation}
    \h = (\hpose, \hidentity, \hglobal),
\end{equation}
\vspace{-1.5pt}
where $\hglobal \in \mathbb{R}^{9}$ is a 6-DoF global pose, and $\hpose \in \mathbb{R}^{51\times3}$ and $\hidentity \in \mathbb{R}^{10}$ are the pose and shape parameters respectively of a template function that maps them to a triangular mesh. In {\methodname} we rely on a decimated version of SMPL with vertices $\htemplatev \in \mathbb{R}^{690}$, reducing the computations while retaining the capability to recover the full template mesh.

For objects, the canonical geometry is given as input by the user and serves as conditioning for our model. We represent it as $\objconditioning = (\objfeatures, \objonehot)$, consisting of $\objfeatures \in \mathbb{R}^{1024}$ PointNeXt~\cite{qian2022pointnext} features and a one-hot class encoding vector $\objonehot$.  Our model diffuses objects' 6-DoF global pose $\objglobal \in \mathbb{R}^9$:
\vspace{-2.5pt}
\begin{equation}
    \obj = (\objglobal).
\end{equation}
\vspace{-1.5pt}

\begin{figure}
    \begin{center}
        \includegraphics[width=0.95\linewidth]{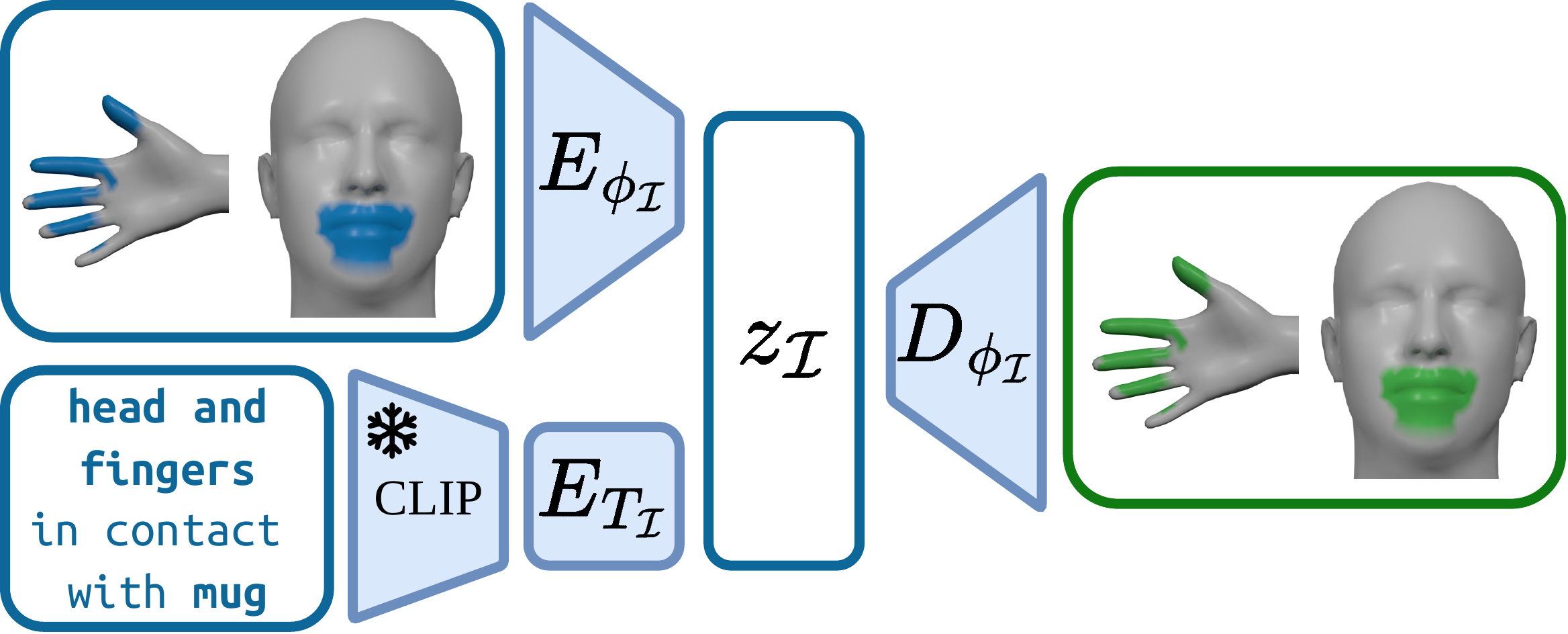}
        \vspace{-4pt}
        \caption{\textbf{Architecture of Contact-Text Interactions model}. We train a mapping from the contact map $E_{\interheat}$ and CLIP embedding $E_{\interlabel}$ to a joint latent space $\interlatent$ that is used to represent the interaction $\inter$. Jointly, we train the decoder $D_{\interheat}$ that maps the latent back to the contact map.}
        \label{fig:contacts}
    \end{center}
\end{figure}
\vspace{-50pt}  %

\begin{figure*}[!ht]
    \vspace{-0.75cm}
    \centering
    \includegraphics[width=0.95\linewidth]{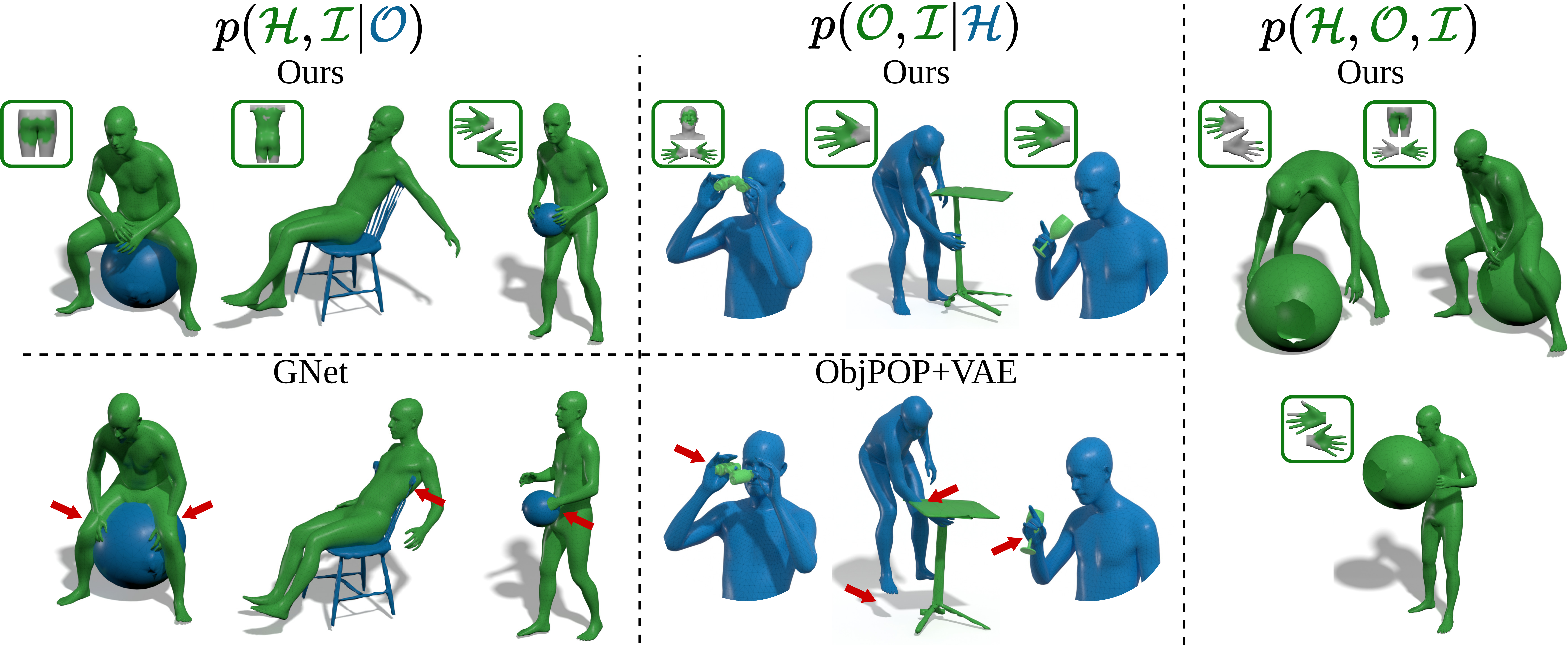}
    \vspace{-0.3cm}
    \caption{\textbf{Comparison with baselines}. 
    In the two left-most columns, we show three samples for 
    $p({\color{pred_col}{\h}},{\color{pred_col}{\inter}} | {\color{input_col}{\obj}})$
    and
    $p({\color{pred_col}{\obj}},{\color{pred_col}{\inter}} | {\color{input_col}{\h}})$
    from BEHAVE and GRAB test sets. 
    {\methodname}'s generations are better aligned with the condition, causing less interpenetration (e.g., for basketball), respecting fine-grained details (e.g., for smaller objects), and demonstrating more diversity for limbs not restricted by contacts (e.g., for yoga ball).
    On the right, {\methodname} is the \emph{only model} that can sample from 
    $p({\color{pred_col}{\h}},{\color{pred_col}{\obj}},{\color{pred_col}{\inter}})$.}
    \label{fig:comparison}
    \vspace{-0.45cm}
\end{figure*}

\paragraph{Interactions.} 
Representing interaction $\inter$ is particularly challenging as we want to combine the intuitiveness of text descriptions with the expressiveness of contact maps.
Our solution is to learn a compact latent representation that encodes both in a joint space.
Given a set of pairs $(\interlabel, \interheat)$, where $\interlabel$ is a text description and $\interheat \in \{0,1\}^{690}$ is a contact map defined on $\htemplatev$, we simultaneously train two encoders $E_{\interheat}(\interheat)=\interlatent \in \mathbb{R}^{128}$ for contact maps and $E_{\interlabel}(\textnormal{CLIP}(\interlabel))=\interlatent$ for CLIP embedding of $\interlabel$, as well as decoder $D_{\interheat}$ mapping the latent space back to the contact map $\interheat$.
We optimize them with the following loss:

\begin{equation}
    \begin{aligned}
    L_{CT}(\interlabel, \interheat)  = &  \mathrm{BCE}(D_{\interheat}(E_{\interheat}(\interheat)),\interheat) + \\ & \mathrm{BCE}(D_{\interheat}(E_{\interlabel}(\interlabel)),\interheat) + \\ & \|E_{\interlabel}(\interlabel) - E_{\interheat}(\interheat)\|_2,
    \end{aligned}
\end{equation}
where $\mathrm{BCE}$ is the Binary-Cross Entropy loss, and the loss terms are auto encoding loss, text-to-contact map encoding loss, and latent space similarity loss. A sketch of this module can be seen in Fig.\ref{fig:contacts}.
Thus, the interactions are represented via a compact code from the unified latent space:

\begin{equation}
    \inter = (\interlatent).
\end{equation}
\noindent %

\subsection{{\methodname}: Trilateral Diffusion for HOI}
\label{sec:method_diffusion}
\paragraph{Diffusion formulation.} To model the joint distribution of Human $\h$, Object $\obj$, and Interaction $\inter$ we formulate a three-way diffusion. For brevity, we define:
\begin{equation}
    \vspace{-1pt}
    \begin{aligned}
    \mathbb{E}_p &\equiv \mathbb{E}_{(\h^0, \obj^0, \mathbf{\inter}^0) \sim p(\h,\obj, \inter)},\\
    \mathbb{E}_t &\equiv \mathbb{E}_{(t^\h, t^\obj, t^\inter) \sim \mathcal{U}\{0,...,T\}^3},\\
    \mathbb{E}_q &\equiv \mathbb{E}_{\h^{t^\h} \sim q(\h^{t^\h} | \h^0),
                                \obj^{t^\obj} \sim q(\obj^{t^\obj} | \obj^0),
                                \inter^{t^\inter} \sim q(\inter^{t^\inter} | \inter^0)}
    \end{aligned}
    \vspace{-1pt}
\end{equation}
Hence, the parameters $\psi$ of a model $\methodnetwork$ are optimized by minimizing the objective (extending Eq.~\ref{eqn:unidiffuser}):
\begin{fleqn}\begin{equation}
    \vspace{-1pt}
    \begin{aligned}
    \min_\psi \mathbb{E}_p \mathbb{E}_t \mathbb{E}_q
        \| & \methodnetwork(
            \h^{t^\h}, \obj^{t^\obj}, \inter^{t^\inter} ; 
            t^\h, t^\obj, t^\inter; \objconditioning)\\
          & - (\h^0, \obj^0, \inter^0) \|_2.
    \label{eqn:trilateraldiffuser}
    \end{aligned}
    \vspace{-1pt}
\end{equation}\end{fleqn}
In practice, we build our method on top of a transformer \cite{vaswani2017attention} architecture with an additional embedding layer for all the input modalities that maps them into a common token space. Formally, $\methodnetwork$ is defined as:
\begin{fleqn}\begin{equation}
    \vspace{-1pt}
    \begin{aligned}
      \methodnetwork: 
        &( \hpose^{\htime},\hidentity^{\htime},\hglobal^{\htime}, 
          \objglobal^{\objtime}, \interlatent^{\intertime}; 
          \htime,\objtime, \intertime, 
          \objconditioning
        )  \mapsto \\
        & (\hat{\h}, \hat{\obj}, \hat{\inter}) \equiv
        (
            {\predhpose},{\predhidentity},{\predhglobal}, 
            {\predobjglobal}, {\predinterlatent}
        ).
    \end{aligned}
    \vspace{-1pt}
\end{equation}\end{fleqn}
We remark that the only required conditioning for {\methodname} is the object representation $\objconditioning$, while other inputs are optional depending on the operating mode. To help the network learn the relation between the different modalities, the triplet $\h$, $\obj$, $\inter$ is tokenized, and we use token-level self-attention to attend fine-grained interaction among the three modalities.

\paragraph{Guidance.} 
Despite explicitly modeling the interaction modality, the diffusion predictions do not always satisfy the contact. 
For 3D HOI, this is a hard constraint to respect in order to avoid floating objects and interpenetrations.
In order to enforce contacts through the denoising process, we adopt a classifier-based guidance \cite{dhariwal2021diffusion} that perturbs the model's prediction on every diffusion step following the feedback of a supervising function $\mathcal{F}$. 

Our idea is to force the human to be in contact with the object where the contact map is active. 
The contact map $\predinterheat=D_{\interheat}(\predinterlatent)$ predicted by {\methodname} enables such use of self-supervised guidance at each diffusion step. We formulate the supervising function as:
\begin{equation}
    \mathcal{F}(\hat{\h}, \hat{\obj}, \hat{\inter}) =
        \sum\limits_{j \in |\htemplatev|} |\predinterheat{_j} \hat{\mathbf{d}}_{j}|,
    \label{eqn:diffusion_inference_guidance}
    \vspace{-6pt}
\end{equation}

where $\hat{\mathbf{d}} \in \mathbb{R}^{690}$ contains for every vertex of the predicted human $\predhtemplatev$ the distance to the closest vertex of the predicted object $\predotemplatev$: 
\vspace{-2pt}
\begin{equation}
    \mathbf{\hat{d}}_j= 
        \min_{i \in |\predotemplatev|} \| \predhtemplatev^j - \predotemplatev^i \|_2.
    \label{eq:dist}
\end{equation}
\vspace{-6pt}

We adopt the reconstruction guidance formulation of \cite{ho2022video}, where the predicted sample $(\hat{\h}, \hat{\obj}, \hat{\inter}) = \methodnetwork(\h, \obj, \inter; \htime, \objtime, \intertime, \objconditioning)$ is directly modified on each denoising step. 
The reconstruction guidance with scale $\lambda$ is thus formulated as:
\vspace{-2pt}
\begin{equation}
    (\hat{\h}, \hat{\obj}, \hat{\inter}) \coloneq 
        (\hat{\h}, \hat{\obj}, \hat{\inter}) - 
        \lambda \nabla_{\h^{t^\h}, \obj^{t^\obj}, \inter^{t^\inter}} 
            \mathcal{F}(\hat{\h}, \hat{\obj}, \hat{\inter}).
\end{equation}
\vspace{-18pt}

\begin{table*}[ht!]
    \vspace{-18pt}
    \setlength{\tabcolsep}{8pt}
    \centering
    \footnotesize
    
    \begin{tabular}{lcccccccc}
     \specialrule{.1em}{.05em}{.05em} 
     \multicolumn{8}{c}{\textbf{BEHAVE}}\\
    
        \multirow{2}{*}{\bf Method} & 
            \multicolumn{3}{c}{\bf \genhi} & & 
            \multicolumn{3}{c}{\bf \genoi} \\
        \cmidrule{2-4} \cmidrule{6-8}
                & 1-NNA $(\to 50)$ & COV$\uparrow$ & MMD$\downarrow$ & \phantom{.} 
                & 1-NNA $(\to 50)$ & COV$\uparrow$ & MMD$\downarrow$ \\
         \cmidrule{2-4} \cmidrule{6-8}
        ObjPOP~\cite{petrov2023popup} + cVAE    &
            -                & -                & -                  & &
            $81.36^{\pm0.2}$ & $35.02^{\pm0.1}$ & $0.329^{\pm0.003}$ \\
        GNet \cite{taheri2022goal}              & 
            $80.01^{\pm0.4}$ & $40.71^{\pm0.4}$ & $1.789^{\pm0.036}$ & &
            -                & -                & -                    \\
        \hline
        \methodsingleOI \textbf{(Ours)}             &
            -                & -                & -                  & &
            $65.06^{\pm0.5}$ & $50.49^{\pm0.1}$ & $\second{0.167^{\pm0.001}}$ \\
        \methodsingleHI \textbf{(Ours)}             &
            $\second{69.51^{\pm0.2}}$ & $\second{46.97^{\pm0.4}}$ & $\second{1.358^{\pm0.010}}$ & &
            -                & -                & -                  \\
        \hline
        {\methodname} \textbf{(Ours)}             & 
            $\first{67.89^{\pm0.3}}$ & $\first{47.81^{\pm0.2}}$ & $\first{1.352^{\pm0.005}}$ & &
            $\first{63.72^{\pm0.3}}$ & $\first{51.71^{\pm0.1}}$ & $\first{0.166^{\pm0.001}}$ \rule{0pt}{2.2ex}\\
            \specialrule{.1em}{.05em}{.05em} 
    \end{tabular}

       \begin{tabular}{lcccccccc}
        \specialrule{.1em}{.05em}{.05em} 
       \multicolumn{8}{c}{\textbf{GRAB}}\\
        \multirow{2}{*}{\bf Method} & 
            \multicolumn{3}{c}{\bf \genhi} & &
            \multicolumn{3}{c}{\bf \genoi} \\
        \cmidrule{2-4} \cmidrule{6-8}
                & 1-NNA $(\to 50)$ & COV$\uparrow$ & MMD$\downarrow$ & \phantom{.} 
                & 1-NNA $(\to 50)$ & COV$\uparrow$ & MMD$\downarrow$ \\
        \cmidrule{2-4} \cmidrule{6-8}      
        ObjPOP~\cite{petrov2023popup} + cVAE    & 
            -                & -                & -                  & &
            $82.09^{\pm0.3}$ & $37.52^{\pm0.8}$ & $0.483^{\pm0.061}$ \\
        GNet \cite{taheri2022goal}              & 
            $89.64^{\pm0.8}$ & $39.33^{\pm1.2}$ & $1.422^{\pm0.087}$ & &
            -                & -                & -                  \\
        \hline
        \methodsingleOI \textbf{(Ours)}             &
            -                & -                & -                  & &
            $\second{66.78^{\pm0.8}}$ & $\second{48.27^{\pm0.1}}$ & $\first{0.252^{\pm0.012}}$ \\
        \methodsingleHI \textbf{(Ours)}             &
            $\first{82.65^{\pm0.1}}$ & $\first{42.87^{\pm0.2}}$ & $\first{0.917^{\pm0.004}}$ & &
            -                & -                & -                  \\
        \hline
        {\methodname} \textbf{(Ours)}                                    & 
            $\second{82.71^{\pm0.5}}$ & $\second{42.76^{\pm0.3}}$ & $\second{0.930^{\pm0.012}}$ &  &
            $\first{65.02^{\pm0.7}}$ & $\first{48.84^{\pm1.2}}$ & $\second{0.268^{\pm0.011}}$ \rule{0pt}{2.2ex}\\
            \specialrule{.1em}{.05em}{.05em}
    \end{tabular}

    \vspace{-5pt}
    \caption{\textbf{Quality of Generated Distribution}. {\methodname} is the only one operating in all the modalities and shows better capability in covering data distribution, improving up to $47$\%.}
    \label{tab:gen_compare}
     \vspace{-15pt}
\end{table*}

\subsection{Training}
\label{sec:method_training}
\paragraph{Contact-Text Labeling.} To train our method, we must collect the contact maps $\interheat$ and the text descriptions $\interlabel$, which are unavailable for many 3D HOI datasets. We define a simple automatic annotation procedure: for every training sample, we obtain the human-object distances $\mathbf{d}$ as described in Equation \ref{eq:dist} and threshold them following~\cite{bhatnagar2022behave} to obtain a binary contact map $\interheat$. 
We detect which of the $24$ body parts of the human template contains at least one vertex in contact, and we use this information to compose the labels following one of the predefined templates, e.g., \texttt{"[parts] are in contact with [object]"}. 

\paragraph{Augmentation.} The lack of interaction variability in the datasets has been one of the main challenges for us. Often, data is statistically biased toward right-handed interactions. Surprisingly, no previous human-object interaction modeling method addressed this problem. Hence, we mirror every sample through the ZY plane, doubling the training data. While the lack of perfect symmetry causes small artifacts, we demonstrate that these are negligible, and the augmentation is highly beneficial for generalization.

\paragraph{Losses.} During training, $\methodnetwork$ takes as input the object class condition $\objconditioning$, three timesteps $(t^\h, t^\obj, t^\inter)$, and a noisy version of the tokenized representation of human $\hpose^{t^\h}$, $\hidentity^{t^\h}$ and $\hglobal^{t^\h}$, object $\objglobal^{t^\obj}$, and interaction $\interlatent^{t^\inter}$, generating the predictions $\hat{\hpose}, \hat{\hidentity}, \hat{\hglobal}, \hat{\objglobal}, \hat{\interlatent}$. 
The learning is supervised by the ground truth representations $\hpose, {\hidentity}, {\hglobal}, {\objglobal}, {\interlatent}$ and templates vertex positions ${\htemplatev}, {\otemplatev}$. We also incorporate the supervision on distances $\mathbf{d}$, fostering spatial alignment. We report the loss details in Sup. Mat.

\section{Experiments}
\label{sec:experiments}
In this section, we measure the quality of {\methodname} in terms of distribution and spatial consistency. In Section \ref{sec:comparison}, we compare our method with specialized approaches on different modalities. This is particularly challenging, as {\methodname} is designed as a unified framework, not privileging any particular modality. 
In Section \ref{sec:ablations}, we ablate the components of our method, providing insights into their specific contribution. 
Finally, we demonstrate applications arising from {\methodname} in Section \ref{sec:applications}. 
In the Sup. Mat., we include experiments with unseen geometries, further qualitative and quantitative evaluation of {\methodname}, runtime analysis, and a user study to validate the generation quality.

\paragraph{Datasets}
For our comparisons, we train {\methodname} and baselines on the union of BEHAVE \cite{bhatnagar2022behave} and GRAB \cite{taheri2020grab}, following the train-test split provided by Object Pop-up\cite{petrov2023popup}. We also explore the scalability of {\methodname} by extending the training to InterCap \cite{huang2022intercap} and OMOMO \cite{li2023object}. We provide descriptions of these datasets and their sampling in Sup. Mat.

\subsection{Comparison to one-way methods}

\label{sec:comparison}
\paragraph{Metrics.} We evaluate the \emph{quality of the generated distributions} using three metrics and comparing the generated samples $g \in S_g$ with the reference ones $r \in S_r$ coming from the GT (with $|S_g|=|S_r|$), reporting the statistics across \emph{three} sampling runs.
The Coverage (\textbf{COV})~\cite{achlioptas2018learning} that measures the percentage of samples in $S_r$ that are matched with at least one sample from $S_g$ ($100$ indicates perfect overlap). The Minimum Matching Distance (\textbf{MMD})~\cite{achlioptas2018learning} measures the average distance of samples in $S_r$ to their closest neighbors in $S_g$, quantifying the misalignment of the distributions. The 1-nearest neighbor accuracy (\textbf{1-NNA})~\cite{yang2019pointflow} measures the leave-one-out accuracy over the union of $S_r \cup S_g$; the optimal value is $50$, and it comprehensively assesses the quality of the distribution. 
To evaluate the \emph{Geometrical Consistency of Generation} of humans, we report the Mean Per Joint Position Error (\textbf{MPJPE}) that measures the error in body joints in cm. We also report its value after applying Procustes Analysis (\textbf{MPJPE-PA}) \cite{gower1975generalized}, alleviating the effect of rotation and scale. For the object we employ the vertex-to-vertex ($\mathbf{E_{v2v}}$) and the object center ($\mathbf{E_{c}}$) errors, together with contact accuracy ($\mathbf{Acc_{cont}}$). For {\methodname}, we measure the error for the contact predicted directly by the method and the one calculated from the generated 3D HOI. We refer to Sup. Mat. for metrics' rigorous definitions.

\paragraph{Baselines.} {\methodname} is the first approach \emph{trained only once} and modeling all the seven static HOI combinations. We compare with single-frame methods specialized in different modalities, posing a challenge to our general setup.
We present a comparison with baselines in two modes, {\genhi} and {\genoi}, as those are the most common ones and the only two that have existing methods working with static HOI.
We present the further evaluation of other operating modes in Sup. Mat.
For {\genhi} we rely on \textbf{GNet}~\cite{taheri2022goal}, while for {\genoi} we choose state-of-the-art Object Pop-up~\cite{petrov2023popup} (\textbf{ObjPOP}). 
Since this latter is a regressive method, we provide a further baseline by substituting its object's center MLP with a cVAE (\textbf{ObjPOP}~\cite{petrov2023popup}\textbf{+cVAE}). 
As in \cite{petrov2023popup}, we integrate a Nearest-Neighbor baseline (\textbf{NN}), which uses the condition modality to retrieve the closest example from the training set. For objects, the similarity is in terms of $\objglobal$, and for humans, it is the distance of root centered joints. 
Additionally, to evaluate the benefits of the joint model, we train three variants of {\methodname} that work only in a single mode: \textbf{\methodsingleOI} for {\genoi}, \textbf{\methodsingleHI} for {\genhi}, and \textbf{\methodsingleHO} for {\genho} (results in Sup. Mat.).

\paragraph{Comparison: Quality of Generated Distribution.} For every method, we generate \emph{three} samples for every example from the test sets of BEHAVE~\cite{bhatnagar2022behave} and GRAB~\cite{taheri2020grab}, and we compare the generated and the GT distributions. 
We consider cases when the condition modality is the object {\genhi} or the human {\genoi}. 
We exclude NN and ObjPOP from this comparison since they do not have variance in prediction. In Tab.\ref{tab:gen_compare}, we report the metrics' mean and variance. Despite being more general, {\methodname} outperforms the specialized baselines on all the metrics, improving up to \emph{47\%}. 
The consistently higher COV and lower MMD indicate better diversity of the generated samples, while a 1-NNA close to $50$ suggests the generated samples are non-trivial.
We report a qualitative comparison in Fig.\ref{fig:comparison}. Notably, {\methodname} performs better or on par with \methodsingleHI and \methodsingleOI, indicating that joint training benefits the generalization of the model. To compare the methods further, we conducted a user study that collected 40 responses. In summary, our method's output is frequently preferred w.r.t. the baselines' ones ($\sim$ 89\% of the cases) and on par with the GT samples ($\sim$ 52\% of the cases); further details are in Sup. Mat.

\begin{table}[t!]
    \setlength{\tabcolsep}{1.1pt}
    \centering
    \scriptsize
    
    \begin{tabular}{lccccccc}
    \specialrule{.1em}{.05em}{.05em} 
    \multicolumn{8}{c}{\textbf{BEHAVE}}\\
        \multirow{2}{*}{\bf Method} & 
            \multicolumn{3}{c}{\bf \genhi} & &
            \multicolumn{3}{c}{\bf \genoi}  \\
        \cmidrule{2-4} \cmidrule{6-8}
            & {\tiny MPJPE$\downarrow$} & {\tiny MPJPE-PA$\downarrow$} & $Acc_{cont}\uparrow$ & & \phantom{.} 
            $E_{v2v}\downarrow$ & $E_{c}\downarrow$ & $Acc_{cont}\uparrow$ \\
        NN                                   & 
            30.5 & 14.2 & $\nicefrac{95.0}{\textnormal{NA}}$ & &
            33.2 & 22.0 & $\nicefrac{95.4}{\textnormal{NA}}$ \\
        ObjPOP \cite{petrov2023popup}        & 
            -      & -      & -     & &
            $\first{27.5}$ & 22.6 & $\nicefrac{95.2}{\textnormal{NA}}$ \\
        ObjPOP \cite{petrov2023popup} + cVAE & 
            -      & -      & -     & &
            35.2 & 23.5 & $\nicefrac{93.6}{\textnormal{NA}}$ \\
        GNet \cite{taheri2022goal}           & 
            35.6 & 14.6 & $\nicefrac{94.6}{\textnormal{NA}}$ & &
            -      & -      & -     \\

        \hline
        \methodsingleOI \textbf{(Ours)}             &
            -                & -                & -                  & &
            $\second{27.9}$ & $\second{15.6}$ & $\nicefrac{\second{95.8}}{\first{96.2}}$ \\
        \methodsingleHI \textbf{(Ours)}             &
            $\second{21.0}$ & $\second{12.5}$ & $\nicefrac{\first{95.6}}{\first{96.5}}$ & &
            -                & -                & -                  \\
        \hline
        
        {\methodname} \textbf{(Ours)}                                  & 
            $\first{20.8}$ & $\first{12.3}$ & $\nicefrac{\second{95.5}}{\first{96.5}}$ & &
            $28.0$ & $\first{15.3}$ & $\nicefrac{\first{95.9}}{\second{96.1}}$ \\
            \specialrule{.1em}{.05em}{.05em} 
    \end{tabular}

    \begin{tabular}{lccccccc}
    \specialrule{.1em}{.05em}{.05em} 
    \multicolumn{8}{c}{\textbf{GRAB}}\\
    
        \multirow{2}{*}{\bf Method} & 
            \multicolumn{3}{c}{\bf \genhi} & &
            \multicolumn{3}{c}{\bf \genoi}  \\
        \cmidrule{2-4} \cmidrule{6-8}
            & {\tiny MPJPE$\downarrow$} & {\tiny MPJPE-PA$\downarrow$} & $Acc_{cont}\uparrow$ & & \phantom{.} 
            $E_{v2v}\downarrow$ & $E_{c}\downarrow$ & $Acc_{cont}\uparrow$ \\
        NN                                   & 
            18.9 & 13.0 & $\nicefrac{\second{97.1}}{\textnormal{NA}}$ & &
            13.1 & 11.8 & $\nicefrac{97.8}{\textnormal{NA}}$ \\
        ObjPOP \cite{petrov2023popup}        & 
            -      & -      & -          & &
            \second{9.4} & 7.7 & $\nicefrac{\second{98.1}}{\textnormal{NA}}$ \\
        ObjPOP \cite{petrov2023popup} + cVAE & 
            -      & -      & -          & &
            13.6 & 12.3 & $\nicefrac{97.4}{\textnormal{NA}}$ \\
        GNet \cite{taheri2022goal}           & 
            26.7 & 15.5 & $\nicefrac{96.6}{\textnormal{NA}}$ & &
            -      & -      & -          \\

        \hline
        \methodsingleOI \textbf{(Ours)}             &
            -                & -                & -                  & &
            $\first{6.9}$ & $\first{4.9}$ & $\nicefrac{\first{99.0}}{\first{98.7}}$ \\
        \methodsingleHI \textbf{(Ours)}             &
            $\second{16.0}$ & $\second{11.6}$ & $\nicefrac{\first{98.0}}{\second{98.1}}$ & &
            -                & -                & -                  \\
        \hline
        
        {\methodname} \textbf{(Ours)}                                & 
            $\first{15.3}$ & $\first{11.1}$ & $\nicefrac{\first{98.0}}{\first{98.3}}$ & &
            $\first{6.9}$ & $\second{5.0}$ & $\nicefrac{\first{99.0}}{\second{98.2}}$ \\
            \specialrule{.1em}{.05em}{.05em} 
    \end{tabular}

    \vspace{-4pt}
    \caption{\textbf{Geometrical Consistency of Generation}. {\methodname} shows a high level of consistency both for human and object predictions. Our contact prediction indicates the networks have also learned to reason based on the interaction modality. For contacts, we show both the accuracy of contacts inferred from $\h$ and $\obj$ meshes, as well as diffused contacts $\inter$ (when available).}
    \label{tab:rec_compare}
    \vspace{-20pt}
\end{table}

\paragraph{Comparison: Geometrical Consistency of Generation.} Comparing generations' spatial consistency with ground truth is not straightforward since a condition may lead to multiple solutions. Hence, in Tab \ref{tab:rec_compare}, we report the error considering the best out of \emph{three} samples, thus approximating an upper bound on the performance of the methods. 
\begin{wrapfigure}[11]{R}{0.45\linewidth}
    \vspace{-1cm}
    \footnotesize
    \begin{center}
        \begin{overpic}[trim=0cm 0cm 0cm 0cm,clip, width=0.98\linewidth]{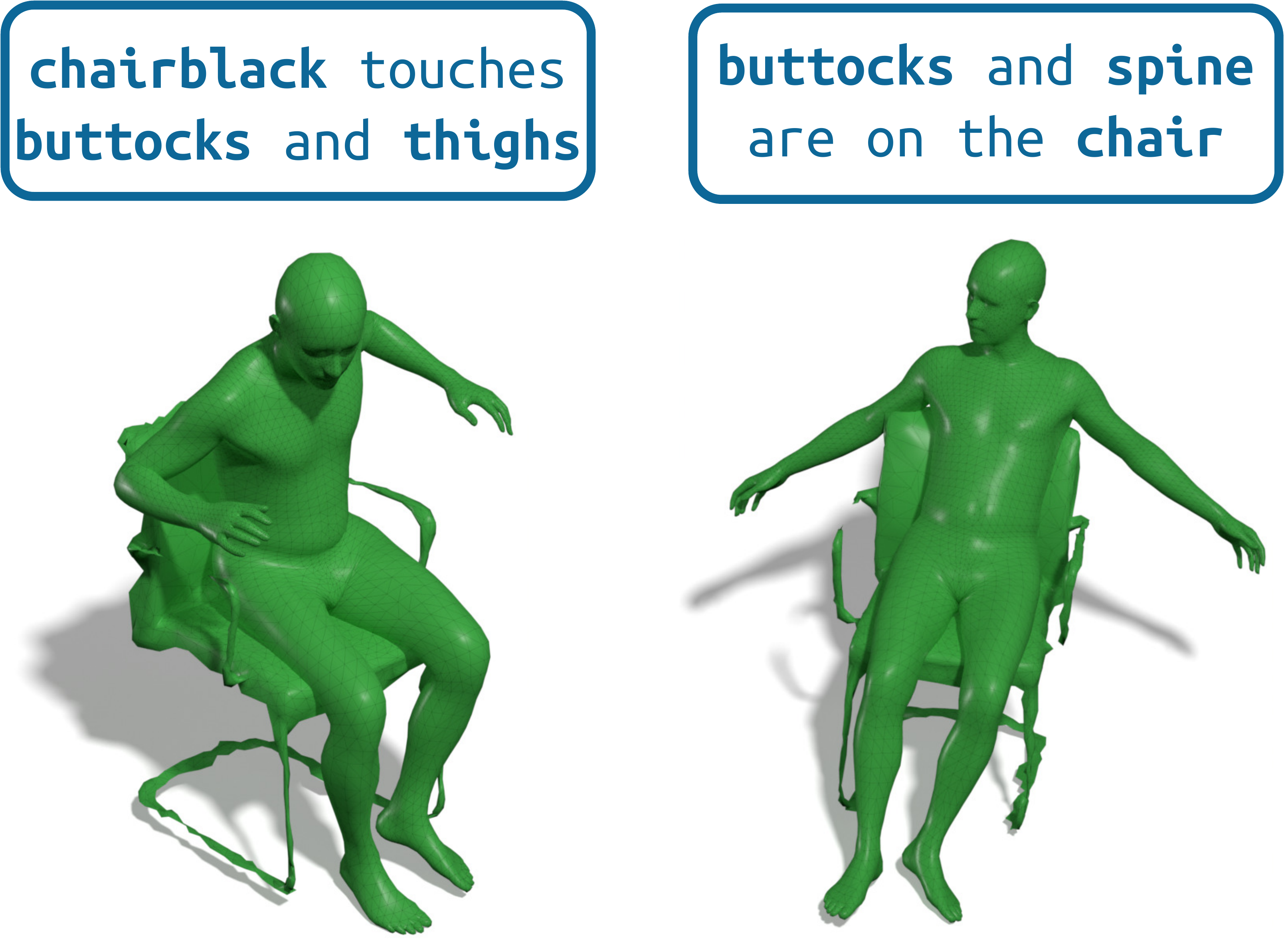}
            \put(39,75){\textcolor{pred_col}{$\obj$},\textcolor{pred_col}{$\h$}$|$\textcolor{input_col}{$\inter$}}
            \put(32,82){\textcolor{pred_col}{\textbf{{\methodname}} (Ours)}}
        \end{overpic}
        \vspace{-0.38cm}
        \caption{\textbf{Text Interaction.} {\methodname} supports text conditioning for $\inter$ modality, providing user control on the contact.}
        \label{fig:comparison_clip}
    \end{center}
\end{wrapfigure}
The improvement over all the errors indicates that our better distribution representation comes with high precision and an understanding of spatial relations. For GRAB, we notice a drastic improvement in object center and orientation, even over the regressive ObjPOP. Considering that our contact is always more accurate, we conclude that our predictions produce more realistic samples. We notice that the predicted contact is better or on par with the contact inferred from meshes, suggesting the network has developed an understanding of the $\inter$ modality. We show in Fig.~\ref{fig:comparison_clip} the flexibility of our $\inter$ representation using varied text descriptions.

\begin{figure*}[t!]
    \vspace{-0.55cm}
    \centering
    \includegraphics[width=0.95\linewidth]{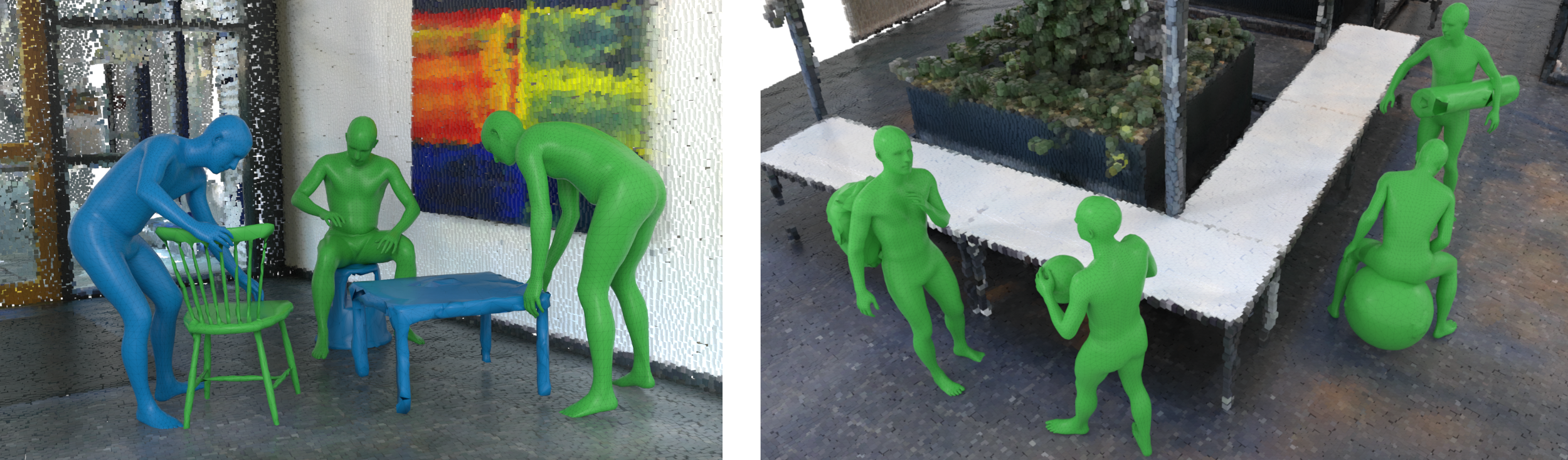}
    \vspace{-0.3cm}
    \caption{\textbf{Scene populating}. Using 3D scans from HPS \cite{guzov2021human}, we validate the practicality of {\methodname} for scene population in various conditioning cases. On the left, we demonstrate conditional synthesis of human-object interactions. On the right, {\methodname} is used for the joint generation of humans and objects.}
    \label{fig:demo_scenes}
    \vspace{-0.35cm}
\end{figure*}

\begin{figure}[ht!]
    \centering
    \includegraphics[trim=0cm 0cm 0cm 0cm,clip,width=\linewidth]{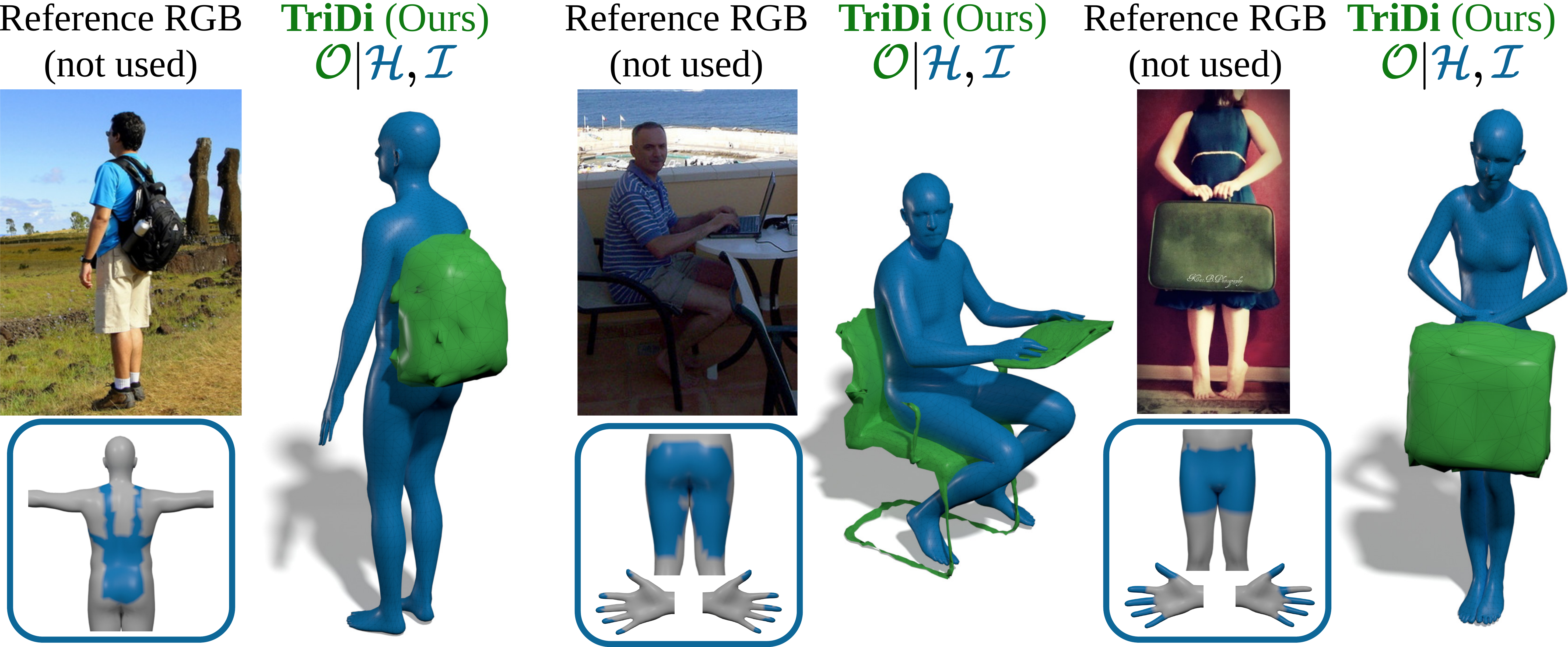}
    \caption{\textbf{Interaction reconstruction}. DECO~\cite{tripathi2023deco} annotates \textcolor{input_col}{human $\h$} and \textcolor{input_col}{contact $\inter$} for the RGB image, while our {\methodname} recovers the \textcolor{pred_col}{object $\obj$}, showing generalization on unseen data distributions.}
    \vspace{-0.4cm}
    \label{fig:demo_deco}
\end{figure}

\subsection{Ablations} 
\label{sec:ablations}
We perform an ablation to analyze the role of the augmentation, $\inter$ modality modeling, and the guidance (we report quantitative results in Sup. Mat.).
The full model obtains the best performance in most metrics. Our augmentation has a valuable effect in improving the 1-NNA, suggesting a better distribution. The guidance and $\inter$ modality plays a crucial role in geometrical consistency, both for the human and the object. The complexity of considering three modalities instead of two seems, in general, beneficial.

\subsection{Applications}
\label{sec:applications}
In this section, we describe applications in populating scenes, interaction reconstructions, and sequences with keyframing. The results are obtained by the same {\methodname} model evaluated in Section~\ref{sec:comparison}. We show generalization to unseen geometries and report samples from a more powerful model trained on more datasets in the Sup. Mat.,
showing scalability to a variety of objects (e.g., vacuum, umbrella, skateboard) and interactions (e.g., feet).

\paragraph{Populating scenes.} Populating scenes is interesting for several downstream tasks in AR/VR, or for synthetic data generation. Here, we demonstrate one possible way to populate scenes with {\methodname}. We first place a virtual object or a human on the ground of a scene from the HPS dataset\cite{guzov2021human}, and then run {\methodname} to generate the complementary modality. Also, we can directly sample $p(\h,\obj,\inter)$ and populate with both humans and objects. Results are shown in Fig.~\ref{fig:demo_scenes}.

\paragraph{Interaction reconstruction.} Our method can also be used to reconstruct interactions from images indirectly. 
In Fig.\ref{fig:demo_deco}, we provide an example from the DAMON dataset of DECO \cite{tripathi2023deco}. DAMON provides contact annotations for images from HOT \cite{chen2023detecting} along with humans estimated by CLIFF~\cite{li2022cliff}. Remarkably, {\methodname} generalizes to such cases despite not being trained on DECO. More examples are included in Sup. Mat.
The requirement of a known 3D mesh can be relaxed by using recent image-to-mesh reconstruction methods such as SDFit~\cite{antic2024sdfit}.

\paragraph{Sequences with keyframing.} Perhaps the most widespread mechanism to generate sequences in animation is via keyframing.  
Here, we can use {\methodname} to automatically generate the keyframes of interaction and use an interpolation to generate a sequence. 
Given a 4D human sequence at 30 fps, we sample at 2 fps, generating the object and interpolate frames in between using slerp for angles and linear for translations. 
We provide an example in our Sup. Mat. video.

\section{Conclusions}
\label{sec:conclusions}
In this work, we proposed {\methodname}, the first joint model for Human, Object, and Interaction, modeling it as a three-variable joint distribution and handling a total of seven different operation modes.
Such versatility of {\methodname} renders prior works as special use cases of the proposed method.
{\methodname} employs an original Contact-Text Interaction representation that combines the interpretability of text with the guidance from the contact information.
Quantitative comparisons demonstrated the superiority of the proposed method both in terms of distribution quality and spatial consistency, with an improvement up to \emph{47}\% over the baseline methods.
Finally, we demonstrated the applicability of the proposed method to scene population, interaction reconstruction from partial data, and generalization to novel object geometry. 
This paves the way for unified HOI usage in content creation, data generation, and AR/VR in the future.

\paragraph{Limitations and future work.}
There are several exciting avenues for future work. We consider scaling beyond single human and object interaction a promising direction to enable the modeling of realistic social situations with increasing complexity. 
The recent advancements in data capturing~\cite{jiang2024scaling, zhang2024hoim3, kim2024parahome, marin2024nicp} open up a possibility of blending scene and object conditioning, laying the foundation for advanced HOI models. 
As a data-driven method, {\methodname} is sensitive to the skewness of the data distribution, expressing more variety towards frequent objects. 
Although {\methodname} shows generalization to unseen geometries (e.g., chairs and stools), we do not expect it to support objects with significantly novel functionality (e.g., wheelchairs, bicycles, bowling balls).

{
\paragraph{Acknowledgements}
Special thanks to Garvita Tiwari, Nikita Kister, and Xianghui Xie for the helpful discussions. We also thank RVH team for their help with proofreading the manuscript. 
This work is funded by the Deutsche Forschungsgemeinschaft - 409792180 (EmmyNoether Programme, project: Real Virtual Humans) and the German Federal Ministry of Education and Research (BMBF): Tübingen AI Center, FKZ: 01IS18039A. 
G. Pons-Moll is a member of the Machine Learning Cluster of Excellence, EXC number 2064/1 – Project number 390727645. 
The authors thank the International Max Planck Research School for Intelligent Systems (IMPRS-IS) for supporting I.~A.~Petrov. 
R. Marin has been supported by the European Union’s Horizon 2020 research and innovation program under the Marie Skłodowska-Curie grant agreement No 101109330. 
The project was made possible by funding from the Carl Zeiss Foundation.
J. Chibane is a fellow of the Meta Research PhD Fellowship Program - area: AR/VR Human Understanding.
}

{
    \small
    \bibliographystyle{ieeenat_fullname}
    \bibliography{main}
}

\setcounter{figure}{0}
\setcounter{table}{0}
\renewcommand{\thefigure}{S\arabic{figure}}
\renewcommand{\thetable}{S\arabic{table}}
\clearpage
\setcounter{page}{1}
\maketitlesupplementary

\begin{abstract}
    This supplementary material provides summary of notation used in the text in \cref{sec:a_background}. We report further implementation details of {\methodname}, description of text labels annotation, insights on symmetry augmentation, and training losses in \cref{sec:a_implementation}. In \cref{sec:a_res}, we include details on the conducted user study, qualitative results on unseen data, ablation results, qualitative results on GRAB, BEHAVE, OMOMO, and InterCap, as well as extended qualitative an quantitative comparison with the baselines. In \cref{sec:a_broader}, we include a discussion on the broader impacts of our work. Details on all four datasets used in the experiments are summarized in \cref{sec:a_datasets}. \cref{sec:a_refinement} introduces an optional post-processing refinement procedure that increases the realism of the generated interactions. Finally, in \cref{sec:a_metrics}, we provide full definition of the error metrics.
    In the attached video, we show results of the keyframing animation discussed in the main text, as well as additional qualitative examples, and we encourage the reader to look at the video. 
\end{abstract}

\section{Background and Notation}
\label{sec:a_background}
\paragraph{Background.} 
We follow the formulation of Denoising Diffusion Probabilistic Model (DDPM) \cite{ho2020denoising} to obtain a closed-form expression for $\mathbf{z}_t$ given the original sample $\mathbf{z}_0$. Let  $\alpha_i=1 - \beta_i$, $\bar{\alpha_t}=\prod_{i=1}^{t}{\alpha_i}$, and $\epsilon \sim \mathcal{N}(0, \mathbf{I})$:
\begin{equation}
    \begin{aligned}
        q(\mathbf{z}_t | \mathbf{z}_0) & = \mathcal{N}(
          \mathbf{z}_t;
          \sqrt{\bar{\alpha_t}}\mathbf{z}_0,
          (1 - \bar{\alpha}_t) \mathbf{I}
        ), \\
        \mathbf{z}_t & = \sqrt{\bar{\alpha_t}} \mathbf{z}_0 + 
          \sqrt{1 - \bar{\alpha}_t} \epsilon.
    \label{eqn:ddpm}
    \end{aligned}
\end{equation}

An iterative denoising process with denoising network $\mathcal{D}_\psi$ is defined by the following:

\begin{equation}
    \mathbf{z}_{t-1} = \sqrt{\bar{\alpha}_{t - 1}} 
      \mathcal{D}_\psi(\mathbf{z}_t; c, t) + 
      \sqrt{1 - \bar{\alpha}_{t-1}} \epsilon,
    \label{eqn:diffusion_inference}
\end{equation}

where $\hat{\mathbf{z}}_0 = \mathcal{D}_\psi(\mathbf{z}_t; c, t)$.

\paragraph{Notation.} Tab.~\ref{tab:sym_table} defines symbols used in our work.
\begin{table}[h!]
    \centering
    \setlength{\tabcolsep}{3pt}
    \begin{tabular}{lll}
    Symbol                   & Description                                     & Domain \\
    \hline 
    $ \h $                  & Human Modality                             &  $(\hpose , \hidentity, \hglobal)$     \\
    $ \hpose $              &  Human Pose                        & $\mathbb{R}^{51 \times 3}$ \\
    $ \hidentity $          &  Human Identity                 & $\mathbb{R}^{10}$ \\
    $ \htemplatev $          &  Human Template's Vertices  & $\mathbb{R}^{690}$ \\
    $ \hglobal $            &  Human Global Pose in 6-DoF     & $\mathbb{R}^9$ \\
    $ \mathbf{d} $          & Human to Object vertex distance & $\mathbb{R}^{690}$ \\
    \hline
    $ \obj $                & Object Modality                        & $(\objglobal)$           \\
    $\objglobal$            &  Object Global Pose in 6-DoF       & $\mathbb{R}^9$ \\
    $ \objconditioning $   & Object Information for conditioning   & $(\objfeatures, \objonehot)$ \\
    $ \objfeatures $            & PointNext features object           & $\mathbb{R}^{1024}$ \\
    $\objonehot$         & one-hot encoding of the class                           & $\{0,1\}^{40}$ \\
    
    $\otemplatev$         &   Object Template's Vertices  & $\mathbb{R}^{1500}$ \\
    \hline
    $\inter$     & Interaction                              & $(\interlatent)$ \\
    $ \interlabel $        & Interaction Textual Label                        &  text \\
    $\interlatent$           & Interaction latent representation                          &   $\mathbb{R}^{128}$ \\
    $ \interheat $       & Interaction contact map                      &  $\{0,1\}^{690}$\\
    $ E_{\interheat}$ & Interaction Encoder (Contact Map) & $\interheat \mapsto \interlatent$\\
    $ D_{\interheat}$ & Interaction Decoder (Contact Map) & $\interlatent \mapsto \interheat$ \\
    $ E_{\interlabel}$ & Interaction Encoder (Textual Label) & $\interlabel \mapsto \interlatent$ \\
     \\ 
    
    \end{tabular}
    \caption{\label{tab:sym_table}\textbf{Notation Table}. The main notation used in our paper.}
\end{table}

\section{Implementation details}
\label{sec:a_implementation}
 The denoising network has a total of $15$M parameters, and it is trained end-to-end. We use a batch size of $1024$, a learning rate of $1e-4$ with a cosine scheduler, and warm up the training during the first $50$k steps. The parameters are optimized with AdamW \cite{loshchilov2017decoupled}. We train for a total of $300$k steps. All the experiments are performed on a machine with RTX4090 GPU. The training of the model takes approximately $20$ hours. The contact encoder-decoder network with $1.7$M parameters is trained separately for $70$ epochs, converging on the same machine in $\sim 1$ hour. The inference for one example with diffusion guidance takes around $3.07$ seconds. Since {\methodname} works per-frame the  inference can be majorly sped up using batching, e.g. inference time for $1024$ examples in one batch is $38.79$ s. All models are implemented in PyTorch \cite{paszke2019pytorch} framework. Following \cite{zhou2019continuity} we convert all rotations ($\hpose, \hglobal, \objglobal$) to 6-d representations before passing them to the network. We rely on blendify~\cite{guzov2024blendify} for visualization.

We implement diffusion reconstruction guidance within the DDPM pipeline and apply it for the last $200$ out of $1000$ iterations of the denoising process with weight $\lambda=2.0$.

\begin{figure}
    \begin{center}
        \includegraphics[width=0.95\linewidth]{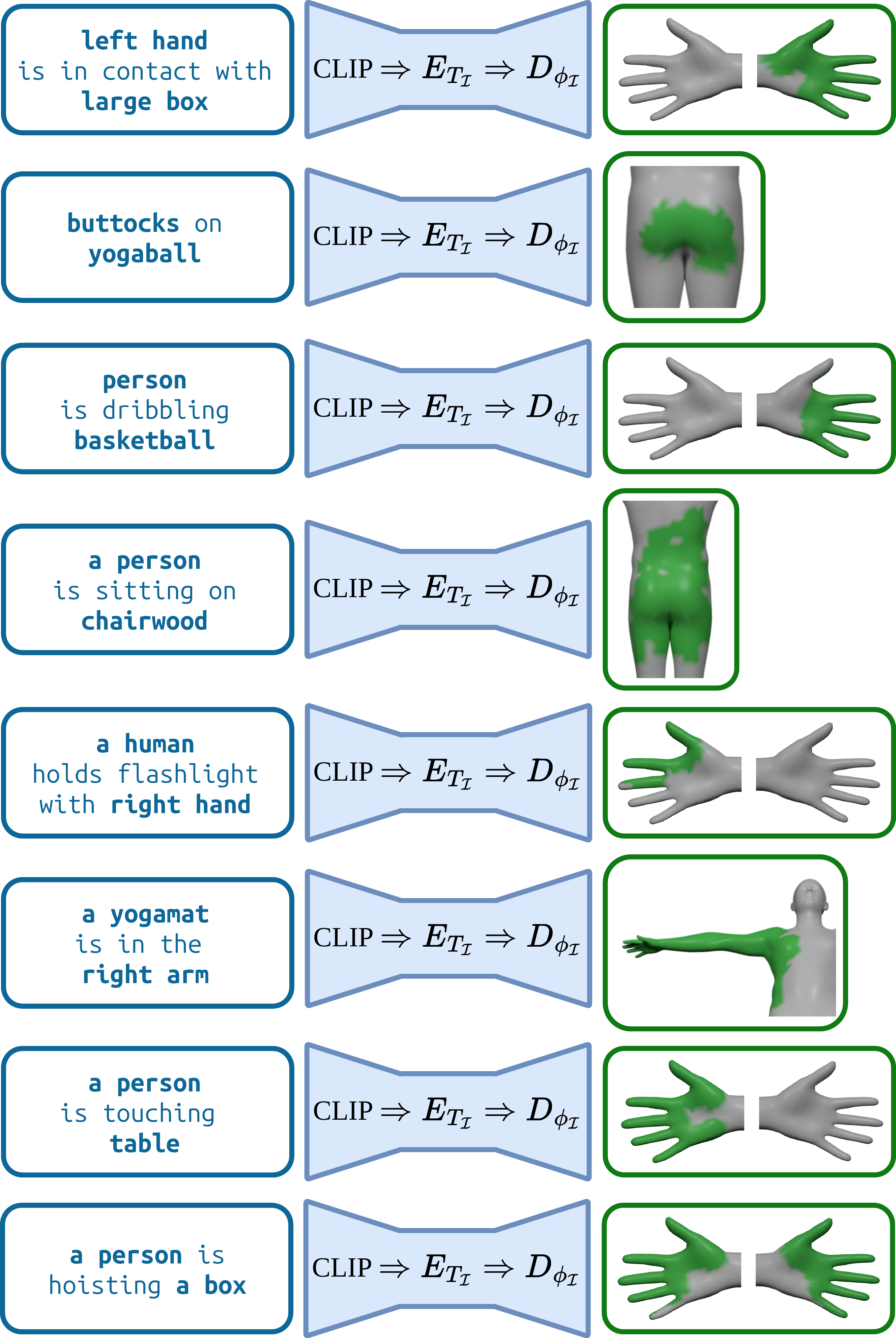}
        \caption{\textbf{Contact maps}.
        Examples of contact maps decoded from text queries.
        }
        \label{fig:sup_contacts}
    \end{center}
\end{figure}

\paragraph{Text labels annotation.} During training, we use a set of predefined templates to generate text labels on the fly, making the encoder $ E_{\interlabel}$ more robust to diverse text inputs.
The template is selected randomly from a pool (provided in Listing~\ref{lst:sup_templates}) based on which body parts are in contact with the object and the object's class. For example, if a person sits on a chair, then the text label is selected from a set of \textit{1. Generic templates} and \textit{2.2 Sitting templates}. 
We study the performance of the contact encoding model in relation to a set of text templates used for training. The model trained using only one generic template (i.e. \texttt{"<body parts> <is / are? in contact with <object class>"}) has a significantly lower recall 63.5 compared to 75.0 of a model trained with the full set of templates. Recall is important because the GT contact maps contain mostly zeros with only a few body points in contact with the object. Moreover, the model trained with the full set of templates exhibits generalization to unseen text inputs (e.g. last row in Fig.~\ref{fig:sup_contacts}).

\paragraph{Augmentation.} During training, we apply the symmetry augmentation randomly mirroring samples through ZY plane. As a result, the model exhibits less bias towards right-handed interactions. Qualitative examples in Fig.~\ref{fig:sup_augmentation} for both cases of sampling from
$p({\color{pred_col}{\h}},{\color{pred_col}{\inter}} | {\color{input_col}{\obj}})$
and
$p({\color{pred_col}{\obj}},{\color{pred_col}{\inter}} | {\color{input_col}{\h}})$
demonstrate how {\methodname} generates left- and right-handed interactions given the same condition.

\begin{verbatimcaption}[ht!]
    \begin{footnotesize}
    \begin{verbatim}
1. Generic templates:
  - <body parts> <is / are> in contact 
    with <object class>
  - <object class> is in contact 
    with <body parts>
  - <body parts> touch(-es) <object class>
  - <object class> <touches> <body parts>

2. Interaction specific templates:
2.1 Basketball template
  - a person is dribbling basketball

2.2 Sitting templates
  - <body parts> <is / are> on <object class>
  - a person <is / sits> on <object class>

2.3 Hands-only templates
  - <object class> is in <body parts>
  - <body parts> <hold(-s) / grab(-s)>
    <object class>
  - a person is <holding / grabbing / carrying> 
    <object class>
    \end{verbatim}
    \end{footnotesize}
    \caption{\textbf{Text labels}. All templates used during training.}
    \vspace{-0.5cm}
    \label{lst:sup_templates}
\end{verbatimcaption}

\paragraph{Losses.} The objective function used to train our network is the weighted combination of the following losses:
\begin{equation}
\begin{aligned}
    L^{\h}_{n}  &= \|\hpose - \predhpose \|_1 + \|\hidentity - \predhidentity \|_1 + \|\hglobal - \predhglobal \|_1 \\
    L^{\obj}_{n}  &= \|\objglobal - \predhglobal \|_1 \\
    L^{\inter}_{n}  &= \|\interlatent - \predinterlatent \|_2 \\
    L^{\h}_{v}  &= \|\htemplatev - \predhtemplatev \|_2 \\
    L^{\obj}_{v}  &= \|\otemplatev - \predotemplatev \|_2 \\
    L^{\inter}_{v}  &= \|\mathbf{d} - \hat{\mathbf{d}} \|_2 
\end{aligned}
\end{equation}
The resulting loss function is:
\begin{equation}
    \begin{aligned}
    L_{{\methodname}} =
        & \lambda^{\h}_{n} L^{\h}_{n} + 
        \lambda^{\obj}_{n} L^{\obj}_{n} + 
        \lambda^{\inter}_{n} L^{\inter}_{n} + \\
        & \lambda^{\h}_{v} L^{\h}_{v} + 
        \lambda^{\obj}_{v} L^{\obj}_{v} + 
        \lambda^{\inter}_{v} L^{\inter}_{v}
    \end{aligned}
\end{equation}
with weighting coefficients set to: $\lambda^{\h}_{n}=\lambda^{\obj}_{v}=2, \lambda^{\obj}_{n}=\lambda^{\inter}_{n}=1, \lambda^{\h}_{v}=6, \lambda^{\inter}_{v}=4$.

\section{Additional Evaluation}
\begin{figure*}[ht!]
    \centering
    \raisebox{-0.5\height}{\begin{minipage}[b]{.45\textwidth}
        \centering
        \includegraphics[width=0.85\linewidth]{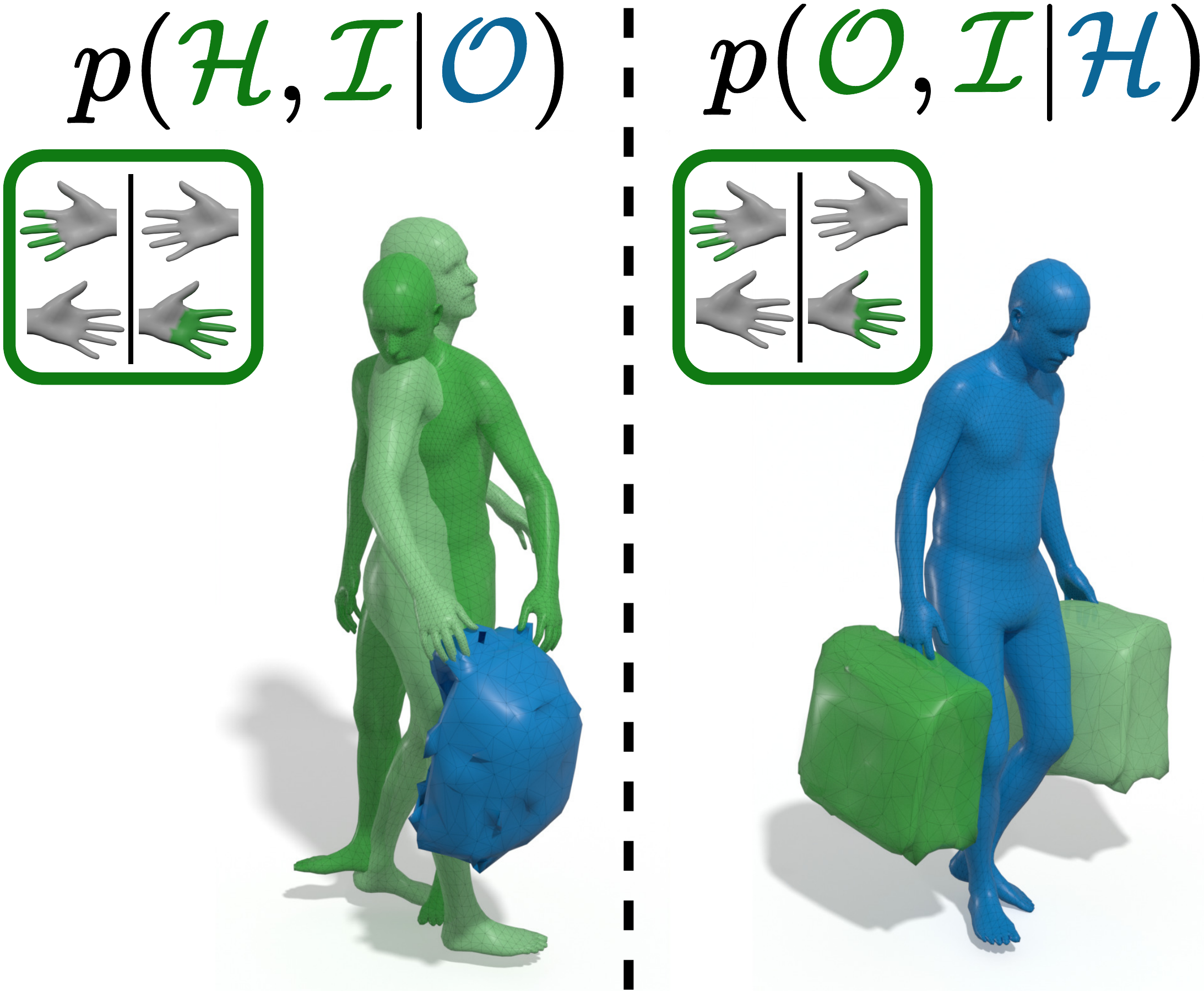}
        \caption{\textbf{Qualitative examples}.
        Results demonstrating the effectiveness of the symmetry augmentation. {\methodname} generates left- and right-handed interactions given the same condition.
        }
        \label{fig:sup_augmentation}
    \end{minipage}}\qquad
    \vspace{-0.4cm}
    \raisebox{-0.5\height}{\begin{minipage}[b]{.48\textwidth}
        \centering
        \includegraphics[width=\linewidth]{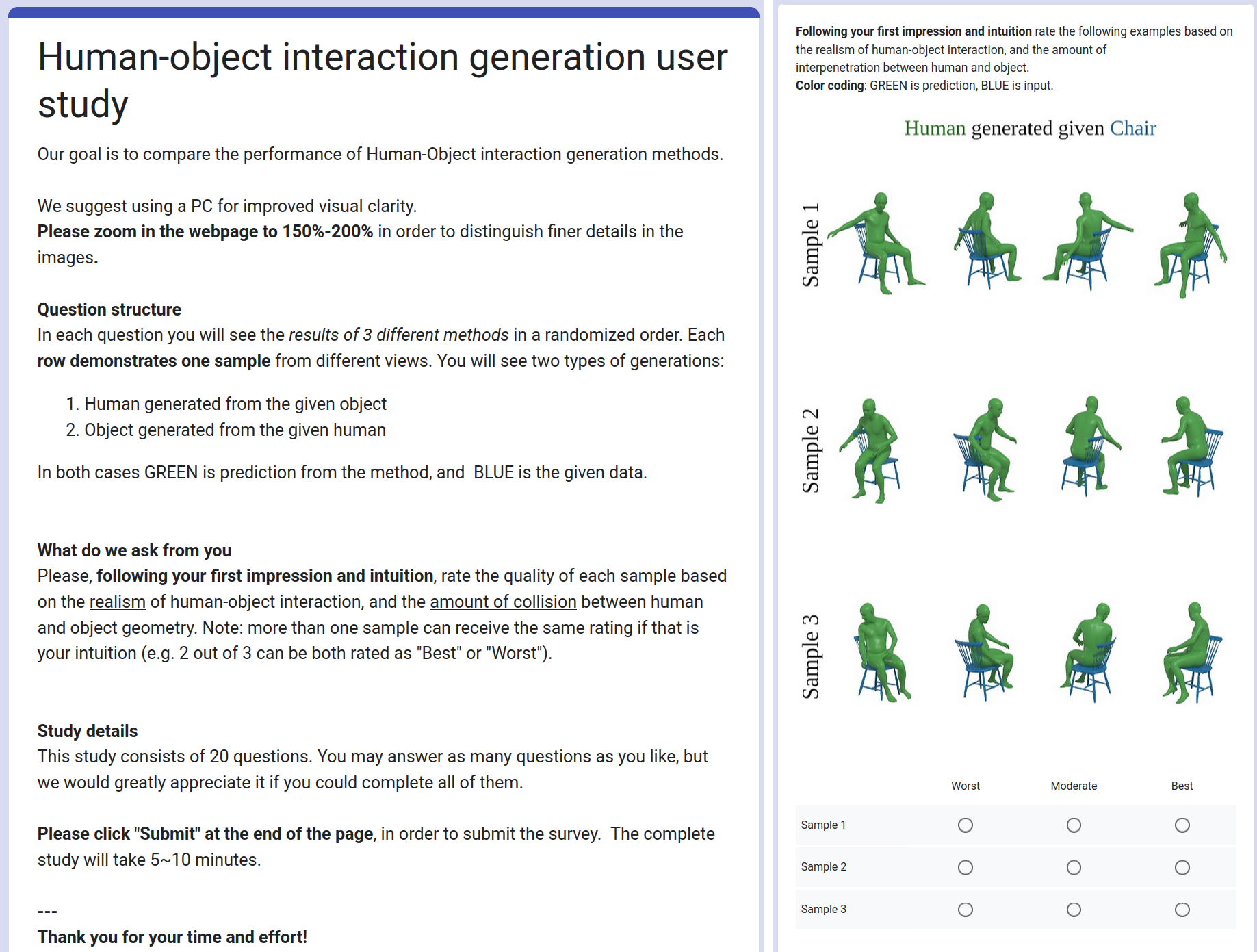}
        \caption{\textbf{User study}.
        The interface of the user study.
        }
        \label{fig:sup_user_study}
    \end{minipage}}
\end{figure*}

\label{sec:a_res}
\paragraph{User study.}
This section introduces details on the user study that was used to evaluate {\methodname}. We have designed and run a user study, asking participants to rate the quality of the generated interactions. We compared {\methodname} against one baseline method and ground-truth data in two generation modes:
$p({\color{pred_col}{\h}},{\color{pred_col}{\inter}} | {\color{input_col}{\obj}})$
and
$p({\color{pred_col}{\obj}},{\color{pred_col}{\inter}} | {\color{input_col}{\h}})$.
We used GNet and ObjPOP+cVAE as the baselines, and randomly selected 10 queries for the generation (5 from each of BEHAVE and GRAB) for each mode. In every question we show users three randomly shuffled samples: ground-truth data, {\methodname}, and corresponding baseline. The participants were asked to rate the quality of each sample based on the realism of human-object interaction, and the amount of interpenetration between human and object. Each sample is rendered from the same 4 orthogonal views to allow comprehensive assessment . The rating scale consisted of three options: \textit{Worst}, \textit{Moderate}, and \textit{Best}, with ratings being non-exclusive (i.e., more than one sample can have a similar rating). Example interface of the user study is provided in the Fig.~\ref{fig:sup_user_study}. As a result, we have collected 40 responses. We summarize the results in the Tab.~\ref{tab:sup_user_study}, comparing the ratings assigned to the samples by users. On average, results of {\methodname} were preferred to the baselines in $89.0\%$ of the cases and preferred to the ground-truth examples in $52.0\%$ of the cases. This suggests that the results of {\methodname} are more appreciable than the baselines and produce a realism comparable to captured data.

\begin{table}[ht!]
    \centering
    \setlength{\tabcolsep}{3pt}
    \begin{tabular}{lll}
        \textbf{Mode} & \textbf{Rating comparison} & \textbf{Result in \%} \\
        \hline
        \multirow{2}{*}{$p({\color{pred_col}{\h}},{\color{pred_col}{\inter}} | {\color{input_col}{\obj}})$} & {\methodname} $>$ GNet & $87.75\%$ \\
         & {\methodname} $>$ GT data & $47.75\%$ \\
        \hline 
        \multirow{2}{*}{$p({\color{pred_col}{\obj}},{\color{pred_col}{\inter}} | {\color{input_col}{\h}})$} & {\methodname} $>$ ObjPOP+cVAE & $90.25\%$ \\
         & {\methodname} $>$ GT data & $56.25\%$ \\
    \end{tabular}
    \caption{\label{tab:sup_user_study}\textbf{User study}. Summary of the user study results.}
    \vspace{-0.5cm}
\end{table}

\paragraph{Evaluation of {\genh}.} We compare {\methodname} with COINS~\cite{zhao2022coins} on the task of human generation given object and text in Table~\ref{tab:sup_coins}. We observe that while COINS is able to generate sitting poses it struggles to generate realistic and diverse interactions with other objects.

\begin{table}[h!]
    \vspace{-5pt}
    \setlength{\tabcolsep}{2.0pt}
    \centering
    \scriptsize
    
    \begin{tabular}{lcccccc}
     \specialrule{.1em}{.05em}{.05em} 
    
        \multirow{3}{*}{\bf Method} & 
            \multicolumn{6}{c}{\bf BEHAVE, \genh} \\
        
        \cmidrule{2-7} 
                & 1-NNA $(\to 50)$ & COV$\uparrow$ & MMD$\downarrow$ & {\tiny MPJPE$\downarrow$} & {\tiny MPJPE-PA$\downarrow$} & $Acc_{cont}\uparrow$ \\
        \cmidrule{2-7}

        COINS             & 
            $96.4^{\pm0.1}$ & $19.6^{\pm0.1}$ & $3.02^{\pm0.008}$ & 43.4 & 15.9 &  $\nicefrac{93.9}{\textnormal{NA}}$ \\
        
        {\methodname}     & 
            $\first{66.1^{\pm0.4}}$ & $\first{50.8^{\pm0.1}}$ & $\first{1.30^{\pm0.010}}$ & $\first{16.9}$ & $\first{10.6}$ & $\nicefrac{\first{96.7}}{\first{99.5}}$\\
        
        \specialrule{.1em}{.05em}{.05em} 
    \end{tabular}
    \vspace{-5pt}
    \caption{\textbf{Comparison with COINS}. }
    \label{tab:sup_coins}
    \vspace{-15pt}
\end{table}

\paragraph{Diversity and multimodality.} We follow Action2Motion~\cite{guo2020action2motion} and compute diversity (Div) and multimodality (MMod) for GT data and {\methodname} to demonstrate that the generated distributions in all \emph{seven} cases are non-trivial. 
Additionally, we evaluate the quality of the generated contacts to prove that the generated HOI is plausible. We compute contact accuracy ($Acc_c$) for cases where GT contacts are available and contact presence ($Presence_c$) that reflects the percentage of generated samples with at least one vertex in contact for other cases. The contact metrics are averaged across three generated samples. The results are presented in Table~\ref{tab:sup_div_mmod}. The variance of the distribution generated by {\methodname} is on par with the variance of the GT data, which means that the generated samples are non-trivial. At the same time, high contact accuracy (96.3 on average) and contact presence (98.4 on average) hint that generated interactions are plausible.
The formulas for Div and MMod are provided in Section~\ref{sec:a_metrics}.
\begin{table*}[t!]
    \setlength{\tabcolsep}{3pt}
    \centering
    \footnotesize
    \begin{tabular}{lcccccccccccccccc}
        \specialrule{.1em}{.05em}{.05em} 
        \multicolumn{16}{c}{\textbf{BEHAVE}}\\
        \multirow{2}{*}{\bf Method} & 
            \multicolumn{3}{c}{\bf \genh} & & 
            \multicolumn{3}{c}{\bf \geno} & & 
            \multicolumn{3}{c}{\bf \geni} & & 
            \multicolumn{3}{c}{\bf \genho} \\
        \cmidrule{2-4} \cmidrule{6-8} \cmidrule{10-12} \cmidrule{14-16}
                & DIV $\to$ & MMod$\to$ & $Acc_{c}\uparrow$ & \phantom{.} 
                & DIV $\to$ & MMod$\to$ & $Acc_{c}\uparrow$ & \phantom{.} 
                & DIV $\to$ & MMod$\to$ & $Acc_{c}\uparrow$ & \phantom{.} 
                & DIV $\to$ & MMod$\to$ & $Acc_{c}\uparrow$ \\
        \cmidrule{2-4} \cmidrule{6-8} \cmidrule{10-12} \cmidrule{14-16}
        GT                      &
            $4.32$ & $4.15$ & - & &
            $2.32$ & $2.20$ & - & &
            $6.68$ & $6.16$ & - & &
            $4.99$ & $4.75$ & - \\
        {\methodname}           & 
            $4.43$ & $4.23$ & $94.5\pm1.2$ & &
            $2.34$ & $2.21$ & $94.8\pm1.2$ & &
            $5.29$ & $4.75$ & $96.2\pm0.1$ & &
            $5.25$ & $4.98$ & $94.9\pm1.5$ \\
        \specialrule{.1em}{.05em}{.05em} 
    \end{tabular}

\vspace{0.1cm}

    \begin{tabular}{lcccccccccccc}
        \specialrule{.1em}{.05em}{.05em} 
        \multicolumn{12}{c}{\textbf{BEHAVE}}\\
        \multirow{2}{*}{\bf Method} & 
            \multicolumn{3}{c}{\bf \genhi} & & 
            \multicolumn{3}{c}{\bf \genoi} & & 
            \multicolumn{3}{c}{\bf \genhoi} \\
        \cmidrule{2-4} \cmidrule{6-8} \cmidrule{10-12}
                & DIV $\to$ & MMod$\to$ & $Presence_{c}\uparrow$ & \phantom{.} 
                & DIV $\to$ & MMod$\to$ & $Presence_{c}\uparrow$ & \phantom{.} 
                & DIV $\to$ & MMod$\to$ & $Presence_{c}\uparrow$ \\
        \cmidrule{2-4} \cmidrule{6-8} \cmidrule{10-12}
        GT                      &
            $8.09$ & $7.55$ & - & &
            $7.15$ & $6.62$ & - & &
            $8.47$ & $7.92$ & - \\
        {\methodname}           & 
            $8.86$ & $8.16$ & $98.8\pm0.1$ & &
            $7.89$ & $7.15$ & $99.3\pm0.1$ & &
            $9.28$ & $8.73$ & $96.1\pm2.2$ \\
        \specialrule{.1em}{.05em}{.05em} 
    \end{tabular}

\vspace{0.3cm}  

    \begin{tabular}{lcccccccccccccccc}
        \specialrule{.1em}{.05em}{.05em} 
        \multicolumn{16}{c}{\textbf{GRAB}}\\
        \multirow{2}{*}{\bf Method} & 
            \multicolumn{3}{c}{\bf \genh} & & 
            \multicolumn{3}{c}{\bf \geno} & & 
            \multicolumn{3}{c}{\bf \geni} & & 
            \multicolumn{3}{c}{\bf \genho} \\
        \cmidrule{2-4} \cmidrule{6-8} \cmidrule{10-12} \cmidrule{14-16}
                & DIV $\to$ & MMod$\to$ & $Acc_{c}\uparrow$ & \phantom{.} 
                & DIV $\to$ & MMod$\to$ & $Acc_{c}\uparrow$ & \phantom{.} 
                & DIV $\to$ & MMod$\to$ & $Acc_{c}\uparrow$ & \phantom{.} 
                & DIV $\to$ & MMod$\to$ & $Acc_{c}\uparrow$ \\
        \cmidrule{2-4} \cmidrule{6-8} \cmidrule{10-12} \cmidrule{14-16}
        GT                      &
            $5.95$ & $5.18$ & - & &
            $2.33$ & $1.53$ & - & &
            $4.33$ & $3.77$ & - & &
            $6.45$ & $5.46$ & - \\
        {\methodname}           & 
            $6.79$ & $5.90$ & $96.7\pm0.8$ & &
            $2.26$ & $1.53$ & $97.5\pm0.7$ & &
            $3.52$ & $3.07$ & $98.2\pm0.1$ & &
            $7.39$ & $6.88$ & $97.7\pm0.8$ \\
        \specialrule{.1em}{.05em}{.05em} 
    \end{tabular}

\vspace{0.1cm}

    \begin{tabular}{lcccccccccccc}
        \specialrule{.1em}{.05em}{.05em} 
        \multicolumn{12}{c}{\textbf{GRAB}}\\
        \multirow{2}{*}{\bf Method} & 
            \multicolumn{3}{c}{\bf \genhi} & & 
            \multicolumn{3}{c}{\bf \genoi} & & 
            \multicolumn{3}{c}{\bf \genhoi} \\
        \cmidrule{2-4} \cmidrule{6-8} \cmidrule{10-12}
                & DIV $\to$ & MMod$\to$ & $Presence_{c}\uparrow$ & \phantom{.} 
                & DIV $\to$ & MMod$\to$ & $Presence_{c}\uparrow$ & \phantom{.} 
                & DIV $\to$ & MMod$\to$ & $Presence_{c}\uparrow$ \\
        \cmidrule{2-4} \cmidrule{6-8} \cmidrule{10-12}
        GT                      &
            $7.43$ & $6.60$ & - & &
            $5.04$ & $4.17$ & - & &
            $7.85$ & $6.81$ & - \\
        {\methodname}           & 
            $8.32$ & $7.68$ & $99.7\pm0.1$ & &
            $5.08$ & $4.39$ & $99.3\pm0.1$ & &
            $9.08$ & $8.61$ & $97.3\pm1.9$ \\
        \specialrule{.1em}{.05em}{.05em} 
    \end{tabular}

    \caption{\textbf{Evaluation of diversity and multi-modality for all sampling modes.} The variance of the distribution generated by {\methodname} is on par with the variance of the GT data, which means that the generated samples are non-trivial. At the same time high contact accuracy (96.3 on average) and contact presence (98.4 on average) hint that generated interactions are plausible.}    
    \label{tab:sup_div_mmod}
\end{table*}

\begin{table}[ht!]
    \setlength{\tabcolsep}{4pt}
    \centering
    \scriptsize
    
    \begin{tabular}{lcccc}
     \specialrule{.1em}{.05em}{.05em} 
     \multicolumn{4}{c}{\textbf{BEHAVE}}\\
    
        \multirow{2}{*}{\bf Method} & 
            \multicolumn{3}{c}{\bf \genho} \\
        \cmidrule{2-4}
                & 1-NNA $(\to 50)$ & COV$\uparrow$ & MMD$\downarrow$ \\
         \cmidrule{2-4} 
        {\methodsingleHO} \textbf{(Ours)} (CM)           &
            $71.75^{\pm0.3}$ & $47.81^{\pm0.5}$ & ${3.15^{\pm0.01}}$ \\
        {\methodsingleHO} \textbf{(Ours)} (Text)           &
            $74.18^{\pm0.2}$ & $46.33{\pm0.2}$ & ${3.21^{\pm0.02}}$ \\
        \hline
        {\methodname} \textbf{(Ours)} (CM)      &
            $\first{70.03^{\pm0.1}}$ & $\first{48.47^{\pm0.1}}$ & $\first{3.07^{\pm0.02}}$ \\
        {\methodname} \textbf{(Ours)} (Text)      &
            $\second{70.14^{\pm0.3}}$ & $\second{48.10^{\pm0.4}}$ & $\second{3.10^{\pm0.02}}$ \\
        \specialrule{.1em}{.05em}{.05em} 
    \end{tabular}

       \begin{tabular}{lcccc}
        \specialrule{.1em}{.05em}{.05em} 
       \multicolumn{4}{c}{\textbf{GRAB}}\\
        \multirow{2}{*}{\bf Method} & 
            \multicolumn{3}{c}{\bf \genho} \\
        \cmidrule{2-4} 
                & 1-NNA $(\to 50)$ & COV$\uparrow$ & MMD$\downarrow$ \\
        \cmidrule{2-4}
        {\methodsingleHO} \textbf{(Ours)} (CM)             &
            $88.81^{\pm0.6}$ & $36.29^{\pm0.6}$ & ${3.10^{\pm0.08}}$ \\
        {\methodsingleHO} \textbf{(Ours)} (Text)           &
            $89.61^{\pm0.3}$ & $34.81{\pm0.2}$ & ${3.28^{\pm0.03}}$ \\
        \hline
        {\methodname} \textbf{(Ours)} (CM)      &
            $\first{87.53^{\pm0.4}}$ & $\second{37.71^{\pm0.1}}$ & $\first{3.05^{\pm0.01}}$ \\
        {\methodname} \textbf{(Ours)} (Text)      &
            $\second{88.56^{\pm0.3}}$ & $\first{37.42^{\pm0.1}}$ & $\second{3.19^{\pm0.02}}$ \\
            \specialrule{.1em}{.05em}{.05em}
    \end{tabular}

    \caption{\textbf{Quality of Generated Distribution for {\genho}}. {\methodname} outperforms {\methodsingleHO} in both sampling from contact maps and text queries. Text provides weaker conditioning than contact maps, thus the resulting distribution exhibits slightly less diversity.}
    \label{tab:sup_gen_ho_compare}
    \vspace{-0.5cm}
\end{table}

\paragraph{Evaluation of {\genho}.} We compare the performance of {\methodname} with a model {\methodsingleHO} that has the same architecture but is trained specifically on {\genho} task (similar to {\methodsingleOI} and {\methodsingleHI} in Tables 1 and 2 of the main paper). We evaluate the methods in two modes: sampling conditioned on contact maps (CM) and conditioned on text query (Text). The results are summarized in Table~\ref{tab:sup_gen_ho_compare}. {\methodname} benefits from joint training on all the tasks together, generating a more diverse and higher quality distribution compared to the model trained specifically on one task. Results also demonstrate that text provides weaker conditioning, the resulting distribution exhibits slightly less diversity compared to the distribution of generations from contact maps. 

We choose {\methodsingleHO} as a baseline because, to the best of our knowledge, there are no existing methods that are able to generate static human-object interaction from text. We attempted to adapt CG-HOI~\cite{diller2024cghoi} to consider static samples instead of motion. However, we observed that the model failed to converge after being adapted to our setting (training on static examples from GRAB and BEHAVE). Our hypothesis is that CG-HOI is designed to work with motion and is initially trained on a relatively small scale dataset (e.g., 500 short motion sequences for BEHAVE), thus generalization to significantly larger data (e.g., 130k static samples for GRAB and BEHAVE) might be too challenging for this model.

\begin{figure*}[!ht]
    \centering
    \includegraphics[width=1.0\linewidth]{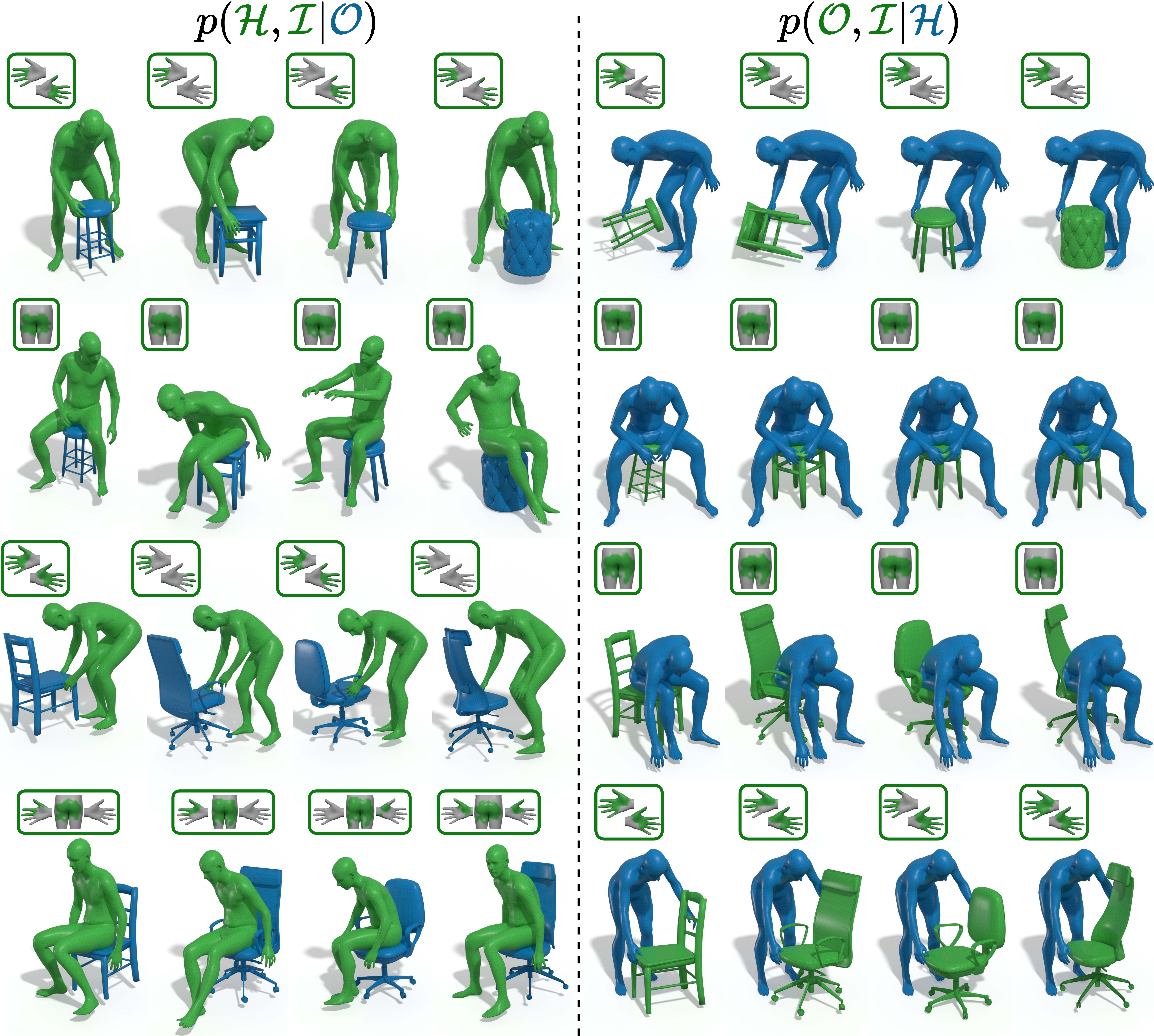}
    \caption{\textbf{Generalization to unseen geometry}. 
    {\methodname} samples from
    $p({\color{pred_col}{\h}},{\color{pred_col}{\inter}} | {\color{input_col}{\obj}})$
    and
    $p({\color{pred_col}{\obj}},{\color{pred_col}{\inter}} | {\color{input_col}{\h}})$
    with unseen objects.
    }
    \label{fig:sup_new_objects}
\end{figure*}

\begin{figure*}[h!]
    \centering
    \includegraphics[width=1.0\linewidth]{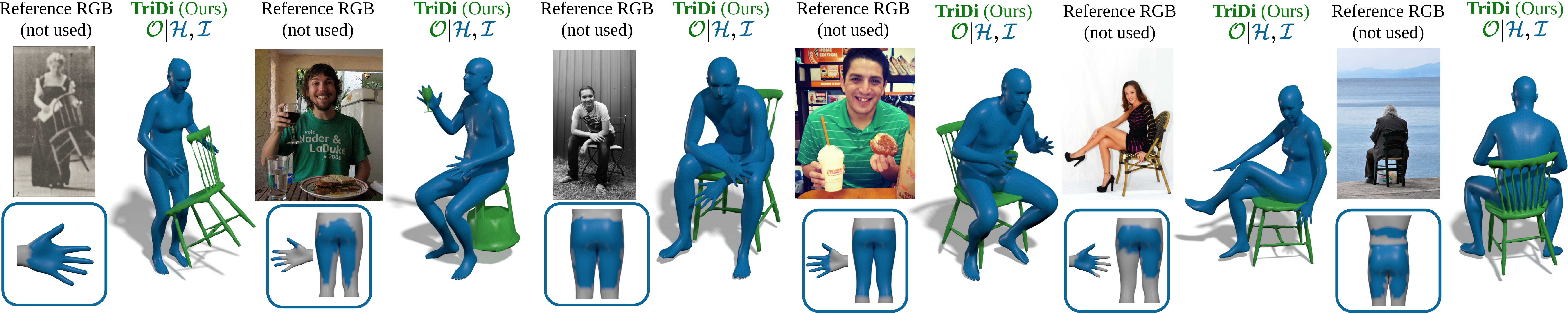}
    \vspace{-0.25cm}
    \caption{\textbf{Interaction reconstruction}. DECO~\cite{tripathi2023deco} annotates \textcolor{input_col}{human $\h$} and \textcolor{input_col}{contact $\inter$} for the RGB image, while our {\methodname} recovers the \textcolor{pred_col}{object $\obj$}, showing generalization on unseen data distributions.}
    \label{fig:sup_deco}
    \vspace{-0.15cm}
\end{figure*}

\paragraph{Generalization to unseen data.} We provide qualitative examples of {\methodname} on eight unseen objects in two sampling modes in Fig.~\ref{fig:sup_new_objects}. The model is able to generate realistic interactions for objects with known functionality.
We also include more examples for interaction reconstruction on the DAMON dataset in Figure \ref{fig:sup_deco}. 

\paragraph{Ablations}
Here, we report the quantitative evaluations of our ablations described in the main paper. Table \ref{tab:gen_abl} covers the quality of the generated distributions, while Table \ref{tab:rec_abl} covers geometrical consistency of the generation.
\begin{table*}[t!]
    \setlength{\tabcolsep}{3pt}
    \centering
    \footnotesize
    \begin{tabular}{lcccccccc}
        \specialrule{.1em}{.05em}{.05em} 
        \multicolumn{8}{c}{\textbf{BEHAVE}}\\
        \multirow{2}{*}{\bf Method} & 
            \multicolumn{3}{c}{\bf \genhi} & &
            \multicolumn{3}{c}{\bf \genoi} \\
        \cmidrule{2-4} \cmidrule{6-8}
                & 1-NNA $(\to 50)$ & COV$\uparrow$ & MMD$\downarrow$ & \phantom{.} 
                & 1-NNA $(\to 50)$ & COV$\uparrow$ & MMD$\downarrow$ \\
        \cmidrule{2-4} \cmidrule{6-8}
        {\methodname}                                    & 
            $\first{67.89^{\pm0.3}}$ & $47.81^{\pm0.2}$ & $\first{1.352^{\pm0.005}}$ & &
            $\first{63.72^{\pm0.3}}$ & $\first{51.71^{\pm0.1}}$ & $\first{0.166^{\pm0.001}}$ \\
         NoGuide                      &
            $\second{68.04^{\pm0.5}}$ & $\first{48.87^{\pm0.2}}$ & $\second{1.355^{\pm0.002}}$ & &
            $\second{63.80^{\pm0.4}}$ & $\second{51.62^{\pm0.3}}$ & $\second{0.167^{\pm0.001}}$ \\
         $(\h, \obj)$                     & 
            $68.19^{\pm0.4}$ & $\second{48.57^{\pm0.1}}$ & $1.373^{\pm0.006}$ & &
            $65.18^{\pm0.5}$ & $50.85^{\pm0.2}$ & $\first{0.166^{\pm0.001}}$ \\
         NoAug                          &
            $69.74^{\pm0.3}$ & $46.21^{\pm0.3}$ & $1.409^{\pm0.009}$ & &
            $69.39^{\pm0.3}$ & $46.20^{\pm0.3}$ & $0.184^{\pm0.002}$ \\
            \specialrule{.1em}{.05em}{.05em} 
    \end{tabular}

\vspace{0.1cm}

    \begin{tabular}{lcccccccc}
        \specialrule{.1em}{.05em}{.05em} 
        \multicolumn{8}{c}{\textbf{GRAB}}\\
        \multirow{2}{*}{\bf Method} & 
            \multicolumn{3}{c}{\bf \genhi} & &
            \multicolumn{3}{c}{\bf \genoi} \\
        \cmidrule{2-4} \cmidrule{6-8}
                & 1-NNA $(\to 50)$ & COV$\uparrow$ & MMD$\downarrow$ & \phantom{.} 
                & 1-NNA $(\to 50)$ & COV$\uparrow$ & MMD$\downarrow$ \\ 
        \cmidrule{2-4} \cmidrule{6-8}
        {\methodname}                                    & 
            $\second{82.71^{\pm0.5}}$ & $\second{42.76^{\pm0.3}}$ & $\second{0.930^{\pm0.012}}$ &  &
            $\first{65.02^{\pm0.7}}$ & $\second{48.84^{\pm1.2}}$ & $\second{0.268^{\pm0.011}}$ \\
        NoGuide                      &
            $82.99^{\pm0.5}$ & $41.74^{\pm1.0}$ & $0.957^{\pm0.007}$ & &
            $\second{65.64^{\pm0.4}}$ & $47.98^{\pm1.3}$ & $0.269^{\pm0.012}$ \\
        $(\h, \obj)$                     &
            $\first{82.40^{\pm1.0}}$ & $42.53^{\pm1.2}$ & $0.996^{\pm0.014}$ & &
            $66.58^{\pm1.7}$ & $\first{49.23^{\pm0.4}}$ & $\first{0.262^{\pm0.002}}$ \\
        NoAug                           &
            $83.05^{\pm1.0}$ & $\first{43.78^{\pm0.6}}$ & $\first{0.878^{\pm0.012}}$ & &
            $67.38^{\pm0.3}$ & $46.11^{\pm0.3}$ & $0.275^{\pm0.006}$ \\
            \specialrule{.1em}{.05em}{.05em}
    \end{tabular}

    \caption{\textbf{Ablation - Quality of Generated Distribution}. Impact of augmentation, $\inter$ diffusion, and guidance.}    
    \label{tab:gen_abl}
\end{table*}

\begin{table}[t!]
\setlength{\tabcolsep}{2pt}
    \centering
    \scriptsize
    \begin{tabular}{lccccccc}
         \specialrule{.1em}{.05em}{.05em} 
     \multicolumn{8}{c}{\textbf{BEHAVE}}\\
        \multirow{2}{*}{\bf Method} & 
            \multicolumn{3}{c}{\bf \genhi} & & 
            \multicolumn{3}{c}{\bf \genoi}  \\
        \cmidrule{2-4} \cmidrule{6-8}
            & MPJPE$\downarrow$ & MPJPE-PA$\downarrow$   & $Acc_{cont}\uparrow$ & &
            $E_{v2v}\downarrow$ & $E_{center}\downarrow$ & $Acc_{cont}\uparrow$ \\       
        \cmidrule{2-4} \cmidrule{6-8}
        {\methodname}                                 & 
            $\first{20.8}$ & $\first{12.3}$ & $\second{95.5} / \first{96.5}$ & &
            $\first{28.0}$ & $\first{15.3}$ & $95.9 / \second{96.1}$ \\
        NoGuide                   &
            $\second{21.5}$ & $\second{12.4}$ & $\first{96.0} / \first{96.5}$ & &
            $\second{28.1}$ & $\second{15.4}$ & $\first{96.2} / \first{96.2}$ \\
        $(\h, \obj)$                   &
            21.9 & 12.7 & $\first{96.0}$ / NA & &
            $28.4$ & 15.6 & $\second{96.1}$ / NA \\
        NoAug                             &
            23.2 & 12.9 & 95.4 / $\second{96.2}$ & &
            31.0 & 17.8 & 95.5 / $96.0$ \\
            \specialrule{.1em}{.05em}{.05em} 
    \end{tabular}

    \begin{tabular}{lcccccccc}
        \specialrule{.1em}{.05em}{.05em} 
       \multicolumn{8}{c}{\textbf{GRAB}}\\
        \multirow{2}{*}{\bf Method} & 
            \multicolumn{3}{c}{\bf \genhi} & &
            \multicolumn{3}{c}{\bf \genoi}  \\
        \cmidrule{2-4} \cmidrule{6-8}
            & MPJPE$\downarrow$ & MPJPE-PA$\downarrow$   & $Acc_{cont}\uparrow$ & &
            $E_{v2v}\downarrow$ & $E_{center}\downarrow$ & $Acc_{cont}\uparrow$ \\     
        \cmidrule{2-4} \cmidrule{6-8}
        {\methodname}                                  & 
            $\second{15.3}$ & $\second{11.1}$ & $\second{98.0} / \second{98.3}$ & &
            $\first{6.9}$ & $\first{5.0}$ & $\first{99.0} / 98.2$ \\
        NoGuide                   &
            $16.2$ & $11.3$ & $97.5$ / $\second{98.3}$ & &
            $9.0$ & $7.5$ & $98.2$ / $\second{98.3}$ \\
        $(\h, \obj)$                   &
            $17.3$ & $11.8$ & $97.3$ / NA & &
            $9.5$ & $7.9$ & $98.0$ / NA \\
        NoAug                         &
            $\first{14.1}$ & $\first{10.4}$ & $\first{98.2 / 98.4}$ & &
            $\second{7.2}$ & $\second{5.2}$ & $\second{98.9}$ / $\first{98.5}$ \\
            \specialrule{.1em}{.05em}{.05em} 
    \end{tabular}

    \caption{\textbf{Ablation - Geometrical Consistency of Generation}. Impact of augmentation, $\inter$ diffusion, and guidance.}
    \label{tab:rec_abl}
\end{table}

\paragraph{Evaluation of penetration.}
We compute SDF-based penetration metrics in Tab.\ref{tab:sup_pen}: average min. dist. between H and O (Min. D. $[cm.]$), contact percentage and penetration score ($C_{\%}, P_{sc} [cm.])$, CHOIS~\cite{zhao2022coins}), penetration depth ($PD [cm.]$, DiffH\textsubscript{2}O~\cite{christen2024diffh2o}).
{\methodname}'s results are close to values computed for ground-truth data, outperforming the baselines that generate floaters and more penetrations.

\begin{table}[h!]
    \vspace{-5pt}
    \setlength{\tabcolsep}{1pt}
    \centering
    \scriptsize
    
    \begin{tabular}{lcccccccccc}
        \specialrule{.1em}{.05em}{.05em} 
        \multicolumn{10}{c}{\textbf{BEHAVE}}\\
    
        \multirow{2}{*}{\bf Method} & 
            \multicolumn{4}{c}{\bf \genhi} & & 
            \multicolumn{4}{c}{\bf \genoi} \\
        \cmidrule{2-5} \cmidrule{7-10}
                & Min. D.$\downarrow$ & $C_{\%}\uparrow$ & $PD\downarrow$ & $P_{sc}\downarrow$ & \phantom{.} 
                & Min. D.$\downarrow$ & $C_{\%}\uparrow$ & $PD\downarrow$ & $P_{sc}\downarrow$ \\
         \cmidrule{2-5} \cmidrule{7-10}
        Data             &
            $\first{0.65}$  & $\first{99.6}$ & $\first{1.73}$ & $\first{0.03}$ & &
            $\second{0.65}$ & $\first{99.6}$ & $\first{1.73}$ & $\first{0.03}$ \\
        \hline
        ObjPOP + cVAE    &
            -    & -    & -   & -  & &
            1.25 & 92.8 & 2.8 & 0.07 \\
        GNet             & 
            1.50 & 91.9 & 3.6 & 0.08 & &
            -    & -    & - \\
        \hline
        {\methodname} \textbf{(Ours)}   & 
            $\second{0.74}$ & $\second{96.5}$ & $\second{2.6}$ & $\second{0.07}$ & &
            $\first{0.56}$  & $\second{97.7}$ & $\second{2.38}$ & $\second{0.05}$ \\
            \specialrule{.1em}{.05em}{.05em} 
    \end{tabular}
    
    \begin{tabular}{lcccccccccc}
        \specialrule{.1em}{.05em}{.05em} 
        \multicolumn{10}{c}{\textbf{GRAB}}\\
       
        \multirow{2}{*}{\bf Method} & 
            \multicolumn{4}{c}{\bf \genhi} & & 
            \multicolumn{4}{c}{\bf \genoi} \\
        \cmidrule{2-5} \cmidrule{7-10}
                & Min. D.$\downarrow$ & $C_{\%}\uparrow$ & $PD\downarrow$ & $P_{sc}\downarrow$ & \phantom{.} 
                & Min. D.$\downarrow$ & $C_{\%}\uparrow$ & $PD\downarrow$ & $P_{sc}\downarrow$ \\
         \cmidrule{2-5} \cmidrule{7-10}     
        Data             &
            $\first{0.08}$ & $\first{100.0}$ & $\second{0.46}$ & $\second{0.0012}$ & &
            $\first{0.08}$ & $\first{100.0}$ & $\second{0.46}$ & $\second{0.0012}$ \\
        \hline
        ObjPOP + cVAE    &
            -    & -    & -    & -    & &
            4.98 & 60.1 & $\first{0.17}$ & $\first{0.0005}$ \\
        GNet             & 
            5.99 & 58.9 & $\first{0.18}$ & $\first{0.0002}$ & &
            -    & -    & - \\
        \hline
        {\methodname} \textbf{(Ours)}   & 
            $\second{0.26}$ & $\second{99.3}$ & 0.79 & 0.0052 & &
            $\second{0.29}$ & $\second{98.8}$ & 0.81 & 0.0048 \\
            \specialrule{.1em}{.05em}{.05em}
    \end{tabular}

    \vspace{-5pt}
    \caption{\textbf{Penetration analysis}.}
    \label{tab:sup_pen}
     \vspace{-5pt}
\end{table}

\begin{figure*}[h!]
    \centering
    \includegraphics[width=1.0\linewidth]{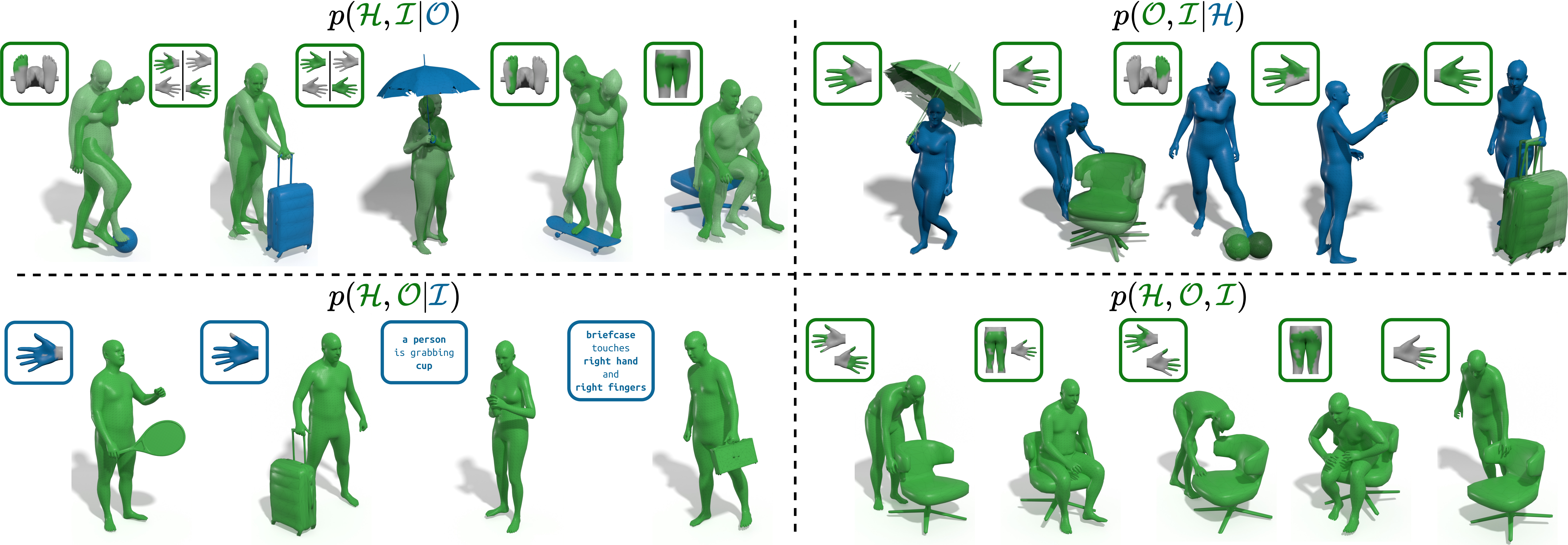}
    \vspace{-0.6cm}
    \caption{\textbf{Qualitative results of {\methodname} on InterCap}.}
    \label{fig:sup_res_intercap}
\end{figure*}

\begin{figure*}[h!]
    \centering
    \includegraphics[width=1.0\linewidth]{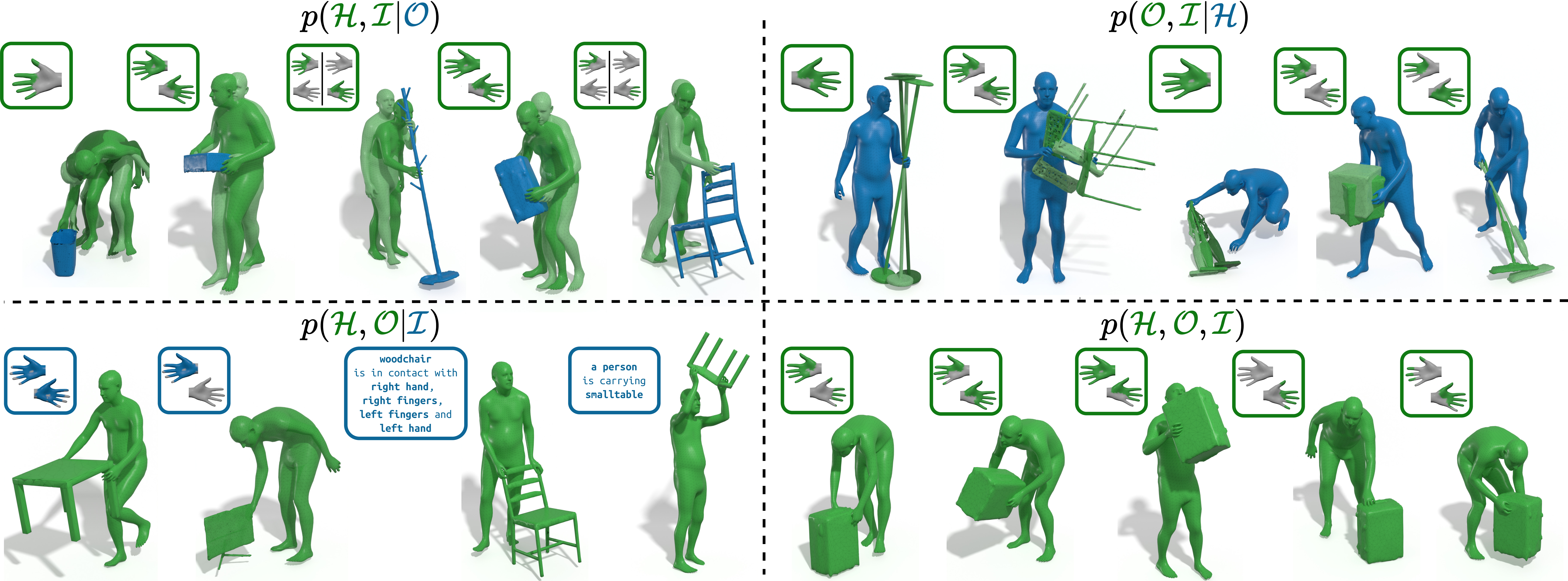}
    \vspace{-0.6cm}
    \caption{\textbf{Qualitative results of {\methodname} on OMOMO}.}
    \label{fig:sup_res_omomo}
\end{figure*}

\begin{figure*}[h!]
    \centering
    \includegraphics[width=1.0\linewidth]{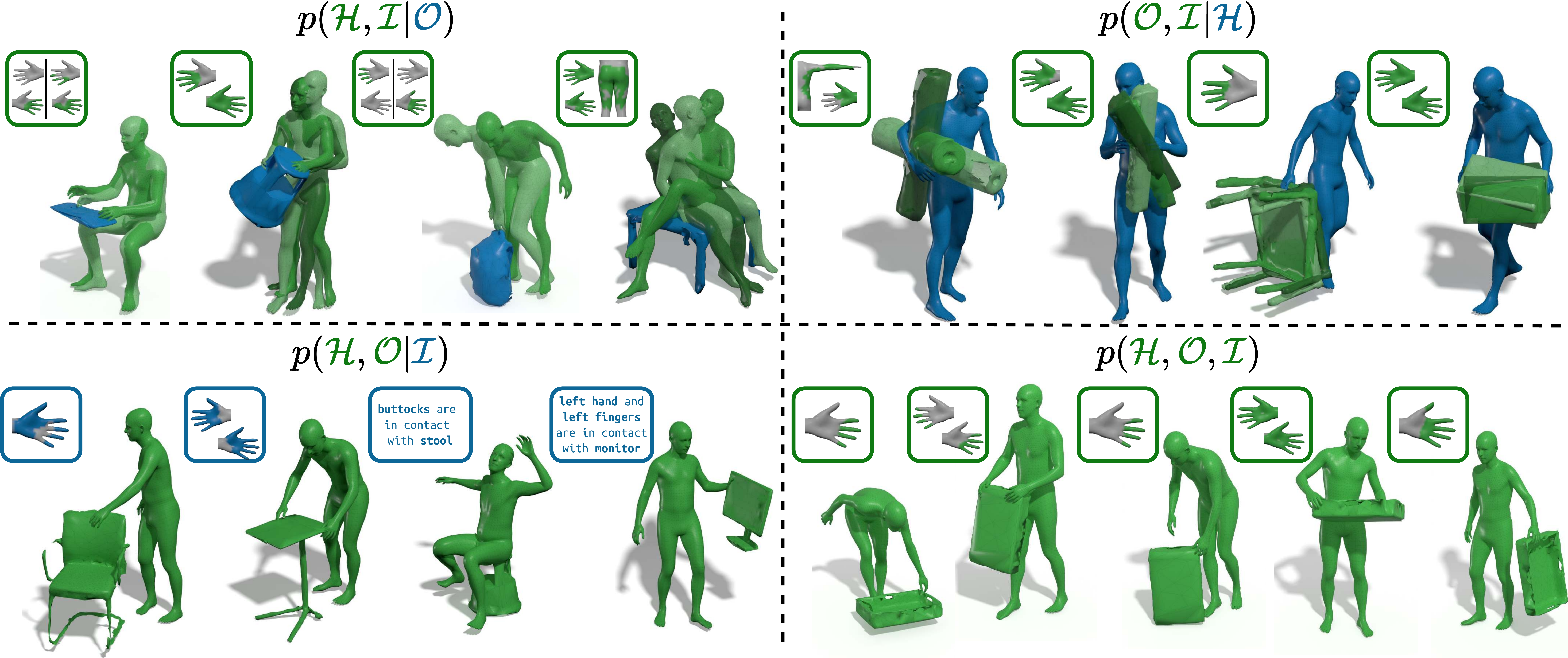}
    \vspace{-0.6cm}
    \caption{\textbf{Qualitative results of {\methodname} on BEHAVE}.}
    \vspace{-0.1cm}
    \label{fig:sup_res_behave}
\end{figure*}

\begin{figure*}[h!]
    \centering
    \includegraphics[width=1.0\linewidth]{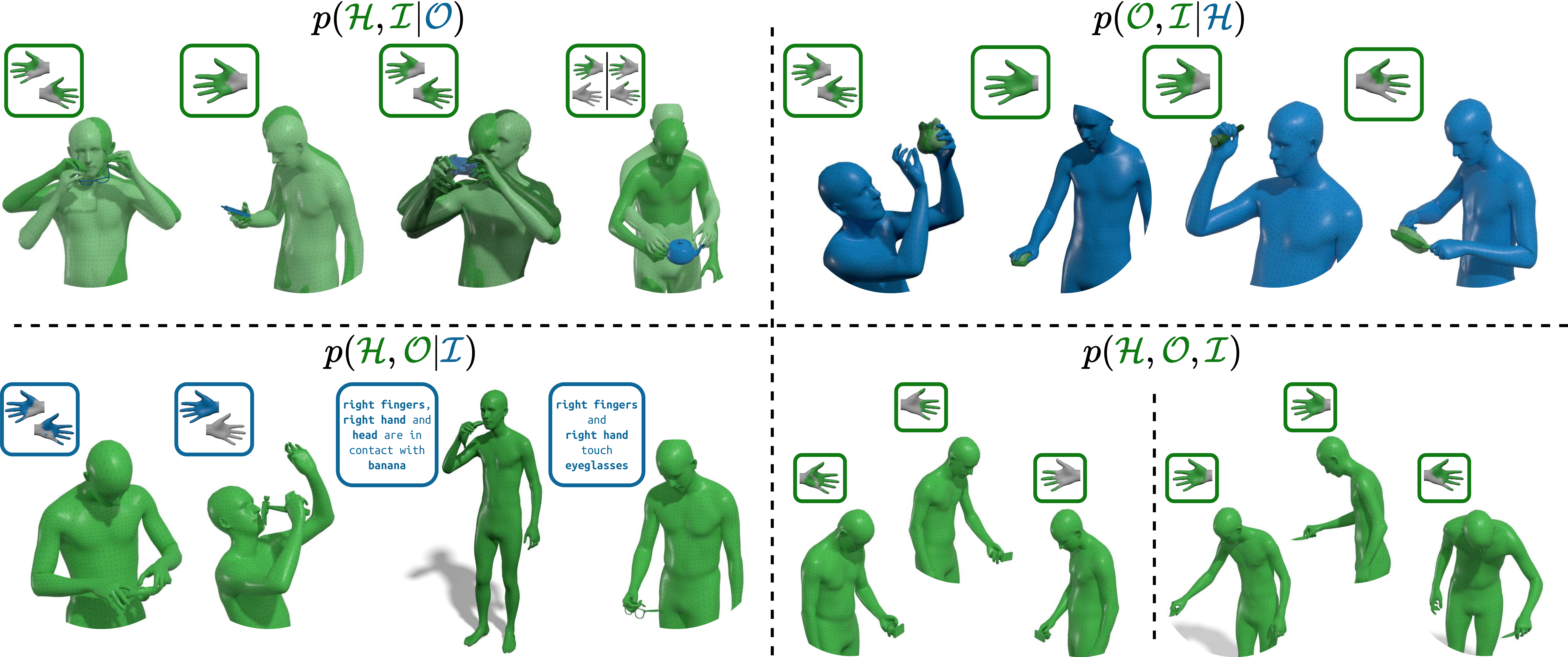}
    \vspace{-0.6cm}
    \caption{\textbf{Qualitative results of {\methodname} on GRAB}.}
    \label{fig:sup_res_grab}
\end{figure*}

\paragraph{Qualitative results}
This section includes additional qualitative results on BEHAVE (Figure \ref{fig:sup_res_behave}) and GRAB (Figure \ref{fig:sup_res_grab}), and introduces examples from InterCap (Figure \ref{fig:sup_res_intercap}) and OMOMO (Figure \ref{fig:sup_res_omomo}).

\begin{figure*}[!ht]
    \centering
    \includegraphics[width=1.0\linewidth]{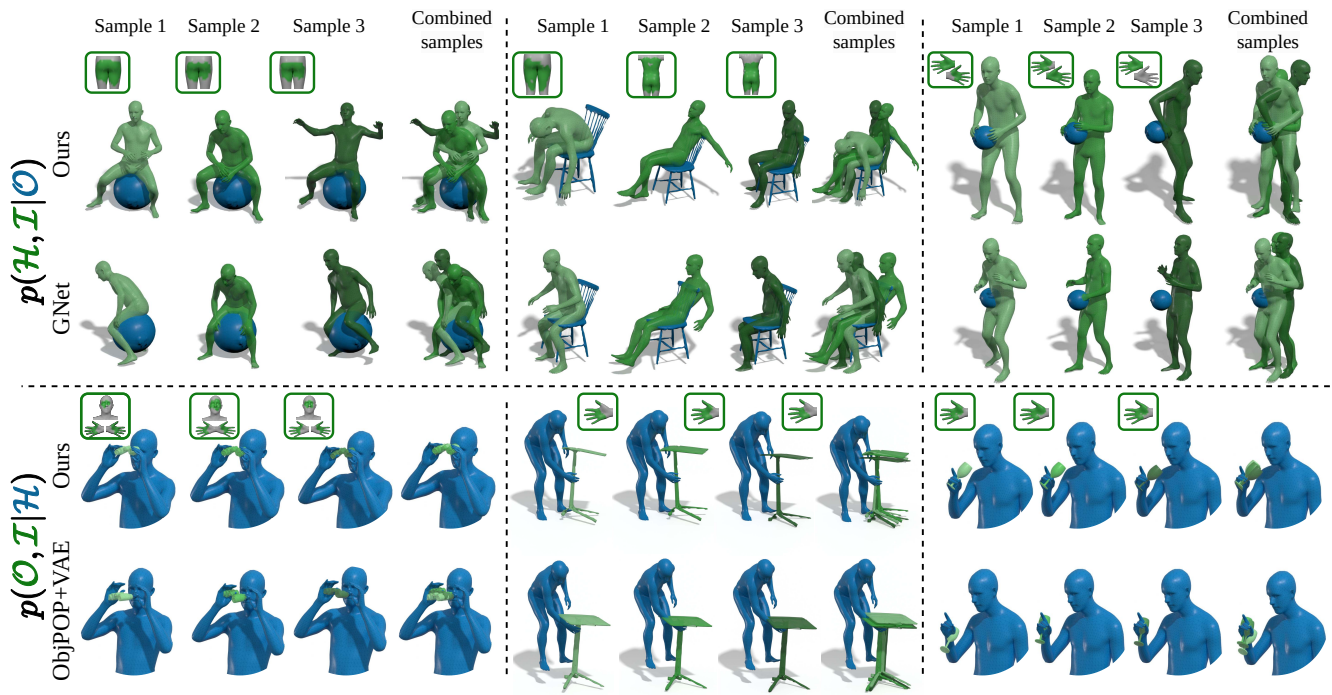}
    \caption{\textbf{Comparison with baselines}. In each group we show three samples (colored in different shades of green) for the same input, as well as one image with the same samples combined. The conditioning is taken from BEHAVE and GRAB test sets.
    }
    \label{fig:sup_comparison}
\end{figure*}

\paragraph{Comparison with baselines} In Fig.~\ref{fig:sup_comparison} we provide an extended comparison with baselines, showing 3 generated samples per same input.

\section{Broader Impacts}
\label{sec:a_broader}
Our method provides an invaluable tool for general content creation and supports analysis of different disciplines like behavioral sciences or ergonomic studies.
Since our method studies human interaction, analysis of subjects' behavior may be included in surveillance applications, leading to privacy issues. However, at the present date, acquiring the 3D data used in our method cannot be easily done without the consensus of the target subject.

\section{Datasets}
\label{sec:a_datasets}
\paragraph{BEHAVE.} BEHAVE \cite{bhatnagar2022behave} captures 8 subjects interacting with 20 different objects, represented as SMPL+H meshes and global configuration, respectively. We downsample the $30 fps$ train sequences to $10 fps$ and consider the official $1 fps$ test subset.
\paragraph{GRAB.} We use the subset of GRAB \cite{taheri2020grab} introduced in \cite{petrov2023popup}. This subset includes 10 subjects interacting with 20 objects. The $120 fps$ train and test sequences are downsampled to $1 fps$. The test set consists of interactions performed by subjects 9 and 10.
\paragraph{InterCap.} We downsample the original $30 fps$ sequences to $10 fps$ and follow the train-test split provided by VisTracker \cite{xie2023visibility}:
Data from subjects 1-8 is used for training, and sequences from subjects 9 and 10 are used for evaluation.
\paragraph{OMOMO.} This dataset captures 17 humans interacting with 15 objects. We employ the official split, using the first 15 subjects for training and subjects 16,17 for testing, and downsample all the sequences to $10 fps$.

\begin{figure}[ht!]
    \begin{center}
        \includegraphics[width=0.95\linewidth]{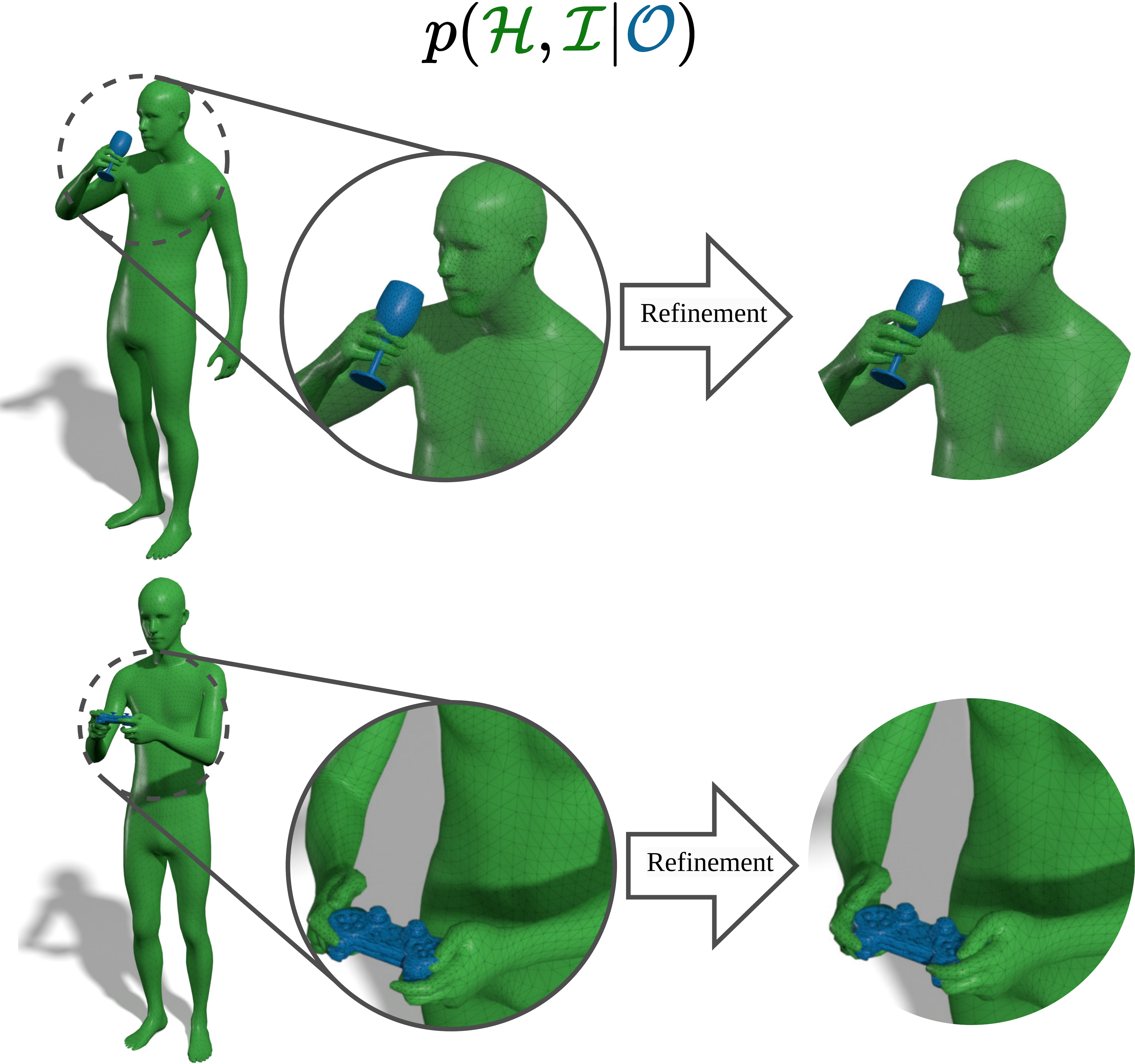}
        \caption{\textbf{Post-processing refinement result}.
        Example results demonstrating the effectiveness of the post-processing refinement. Optionally, {\methodname} results can be refined using an optimization procedure that improves fine hand details.
        }
        \label{fig:sup_refinement}
    \end{center}
\end{figure}
\section{Post-processing refinement}
\label{sec:a_refinement}

\paragraph{Motivation.} In some cases, {\methodname}'s samples may miss perfect plausibility of fine grained details, especially for smaller objects. 
Such behavior is naturally caused by a lack of detailed hand modeling in the majority of the training data. 
To counter this problem, we introduce a post-processing refinement.  
We demonstrate qualitative examples of post-processing refinement in Fig.~\ref{fig:sup_refinement} to show extended capabilities of {\methodname}. The proposed refinement procedure is able to correct mistakes in fine-grained grasps leading to increased realism of predictions. In the following paragraphs we provide details on the post-processing refinement.
We remark that all the qualitative and quantitative results in the main paper and supplementary are obtained without the refinement for a fairer comparison.
\paragraph{Refinement implementation.} We take inspiration from DexGraspNet~\cite{wang2023dexgraspnet} to design an optimization procedure refining the generated hands. The original refinement minimizes the error term:
\begin{equation}
    E_{fc} + w_{dis}E_{dis}+w_{pen}E_{pen}+w_{spen}E_{spen}+w_{prior}E_{prior}
\end{equation}
where $E_{fc}$ is a force closure term proposed in \cite{liu2021synthesizing} that encourages the closed grasp, $E_{dis}$ and $E_{pen}$ are, respectively, attraction and repulsion terms, enforcing contact and penalizing penetration, $E_{spen}$ is a self-penetration term, $E_{prior}$ is a hand prior term penalizing unrealistic pose configurations. We refer to \cite{wang2023dexgraspnet} for detailed definition of the energies. 
We add two more terms to the original energy to adapt the method to our use case. Firstly, we want the final result to don't deviate too much from the initial prediction of {\methodname}, thus we introduce regularization:
\begin{equation}
E_{reg} = \|\hat{\hpose} - \tilde{\hpose}\|_2
\end{equation}
where $\hat{\hpose}$ is human pose predicted by {\methodname} and $\tilde{\hpose}$ is the refined human pose. Secondly, we want to explicitly penalize intersections between hands and objects. To achieve this we introduce a term inspired by \cite{karras2012maximizing, tzionas2016capturing} that detects the collision between hand and object meshes, penalizing the quantity:
\begin{equation}
    \begin{aligned}
    E_{isect} = \sum_{(\mathbf{f_\h}, \mathbf{f_\obj}) \in C} 
        & \left[\sum_{\mathbf{v_\h} \in \mathbf{f_\h}}{\|-\Psi_{\mathbf{f_\obj}}(\mathbf{v_\h})\|^2} \right. + \\
        & \left. \sum_{\mathbf{v_\obj} \in \mathbf{f_\obj}}{\|-\Psi_{\mathbf{f_\h}}(\mathbf{v_\obj})\|^2}
    \right]
    \end{aligned}
\end{equation}
where $\mathbf{v_\h}\in\mathbf{V_\h}$ and $\mathbf{f_\h}\in\mathbf{F_\h}$ are vertices and faces of the human mesh, $\mathbf{v_\obj}\in\mathbf{V_\obj}$ and $\mathbf{f_\obj}\in\mathbf{F_\obj}$ are vertices and faces of the object mesh, $C$ is a set of pairs of collided faces, $\Psi_{\mathbf{f}}: \mathbb{R}^3 \to \mathbb{R}_{+}$ is a cone distance field from the face $\mathbb{f}$ (full definition can be found in \cite{tzionas2016capturing}).

Since {\methodname} deals with full bodies, the optimization procedure is split into two stages: first, to fix the global positioning of the hand (optimization w.r.t. shoulder, elbow, and wrist joints), next to fix the fine details (optimization w.r.t. fingers). Therefore, we obtain the following energy terms:
\begin{equation}
    \begin{aligned}
        E_{stage\_1} = &w_{dis}E_{dis}+w_{pen}E_{pen} + \\
            & w_{reg}E_{reg}+w_{isect}E_{isect} \\
        E_{stage\_2} = &E_{fc} + w_{dis}E_{dis}+w_{pen}E_{pen} + \\
            & w_{spen}E_{spen}+w_{prior}E_{prior} + \\
            & w_{reg}E_{reg}+w_{isect}E_{isect}
    \end{aligned}
\end{equation}
where weights are $w_{dis}=0.2$, $w_{pen}=100$, $w_{reg}=20$, $w_{isect}=400$ for the first stage, and $w_{dis}=w_{pen}=w_{isect}=100$, $w_{spen}=10$, $w_{prior}=0.5$, $w_{reg}=10$ for the second stage. Optimization setup follows~\cite{wang2023dexgraspnet} with 1000 iterations for the first stage and 2000 iterations for the second stage.

\section{Error Metrics}
\label{sec:a_metrics}
\paragraph{Quality of Generated Distribution.} To evaluate our fitting to the target distribution, we use three metrics. The \emph{Coverage (COV)}\cite{achlioptas2018learning}:
\begin{equation}
    COV(S_g,S_r)=\frac{|\{\arg\min\limits_{r \in S_r}D(g,r)|g \in S_g\}|}{|S_r|},
\end{equation}
where $D(g,r)$ is $L2$ distance between corresponding feature vectors, namely, root-centered body joints for humans and concatenated global position and orientation for objects. 

\emph{Minimum Matching Distance} (MMD)\cite{achlioptas2018learning}:
\begin{equation}
    MMD(S_g,S_r)=\frac{1}{|S_r|}\sum_{r \in S_r}\underset{g \in S_g}{\min}D(g,r)
\end{equation}
We employ the same definition of $D(\cdot,\cdot)$ as for COV. 

\emph{1-Nearest Neighbor Accuracy} (1-NNA) \cite{yang2019pointflow}. Given a generated sample $g$, The idea is to evaluate how a 1-NN classifier trained on $S_{-g}=S_r \cup S_g - \{g\}$ would classify the sample $g$. Namely, 1-NNA evaluates the leave-one-out accuracy over the union dataset:
\begin{equation}
    \begin{aligned}
        & 1 \text{-} NNA(S_g,S_r) = \\
        & \frac{\sum_{X \in S_g}\mathds{1}[N_X \in S_g]+\sum_{Y \in S_r}\mathds{1}[N_Y \in S_r]}{|S_g| + |S_r|},
    \end{aligned}
\end{equation}
where $N_X$ is the nearest neighbor of $X$ in $S_{-X}$, $\mathds{1}[\cdot]$ is the indicator function. We define nearest neighbors according to the aforementioned distance metrics $D(\cdot,\cdot)$. 

\emph{Diversity} (Div) \cite{guo2020action2motion}. Diversity measures the variance of the generated samples. Two subsets $S_1=\{v_1,...,v_{|S|}\}$ and $S_2=\{v'_1,...,v'_{|S|}\}$ of the same size $|S|=200$ are drawn from either $S_g$ or $S_r$ (depending on whether we want to evaluate the metric for the method or the GT data). The diversity then is computed as follows:
\begin{equation}
    Div(S_1, S_2) = \frac{1}{|S|}\sum_{i = 1}^{|S|}{\Vert{v_i - v'_i}\Vert_2},
\end{equation}

\emph{Multimodality} (MMod) \cite{guo2020action2motion}. Multimodality measures the variance of the generated samples within the same object category. For every object class $c\in{1,...C}$ two subsets $S^c_1=\{v_{c,1},...,v_{c,|S|}\}$ and $S^c_2=\{v'_{c,1},...,v'_{c,|S|}\}$ of the same size $|S|=200$ are drawn from either $S_g$ or $S_r$. The multimodality is then computed as follows ($S_1=\{S^1_1, ...,S^C_1\}, S_2=\{S^1_2, ...,S^C_2\}$):
\begin{equation}
    MMod(S_1, S_2) = \frac{1}{C * |S|}\sum_{c = 1}^{C}\sum_{i = 1}^{|S|}{\Vert{v_{c,i} - v'_{c,i}}\Vert_2},
\end{equation}

\paragraph{Geometrical Consistency of Generation.} The $E_{v2v}$ error measures the average L2 distance between the position of the predicted object vertices and the ones of the ground truth:
\begin{equation}
E_{v2v} = \frac{1}{|\otemplatev|}\sum\limits_{i \in |\otemplatev|}{\|\otemplatev^i - \predotemplatev^i\|_2}
\end{equation}

The $E_{c}$ error measures the average L2 distance between the position of the predicted object center and the one of the ground truth:
\begin{equation}
    E_{c} = \left\Vert{
        \frac{1}{|\otemplatev|}
            \sum\limits_{i \in |\otemplatev|}{\otemplatev^i} - 
        \frac{1}{|\predotemplatev|}
            \sum\limits_{i \in |\predotemplatev|}\predotemplatev^i
    }\right\Vert_2.
\end{equation}

 We complement the reconstruction metrics with the contact accuracy metric $Acc_{cont}$:
\begin{equation}
  Acc_{cont} = \frac{1}{|\htemplatev|}
        \sum\limits_{i \in |\htemplatev|} 
            \mathds{1}[\predinterheat^i=\interheat^i],
\end{equation}
where $\mathds{1}$ is an indicator function.
\newpage

\clearpage

\end{document}